\relax
\documentclass[letterpaper,11pt]{article} % DO NOT CHANGE THIS
\usepackage{aaai22}  % DO NOT CHANGE THIS
\usepackage{times}  % DO NOT CHANGE THIS
\usepackage{helvet}  % DO NOT CHANGE THIS
\usepackage{courier}  % DO NOT CHANGE THIS
\usepackage[hyphens]{url}  % DO NOT CHANGE THIS
\usepackage{subcaption}
\usepackage{multirow}
\usepackage{graphicx} % DO NOT CHANGE THIS
\usepackage{natbib}  % DO NOT CHANGE THIS AND DO NOT ADD ANY OPTIONS TO IT
\usepackage{caption} % DO NOT CHANGE THIS AND DO NOT ADD ANY OPTIONS TO IT
\usepackage{lineno}
\usepackage{soul}

\DeclareCaptionStyle{ruled}{labelfont=normalfont,labelsep=colon,strut=off} % DO NOT CHANGE THIS
\usepackage{algorithm}
\usepackage{algorithmic}
\usepackage{xcolor}
\usepackage{amsmath}

\usepackage{newfloat}
\usepackage{listings}

\floatstyle{ruled}
\newfloat{listing}{tb}{lst}{}
\floatname{listing}{Listing}
\title{Using AI to Measure Parkinson’s Disease Severity at Home}
\author {
    Md Saiful Islam\textsuperscript{\rm 1}\thanks{Corresponding author. \textbf{Email}: mislam6@ur.rochester.edu},
    Wasifur Rahman\textsuperscript{\rm 1},
    Abdelrahman Abdelkader\textsuperscript{\rm 1},
    Sangwu Lee\textsuperscript{\rm 1},
    Phillip T. Yang\textsuperscript{\rm 2},
    Jennifer L. Purks\textsuperscript{\rm 2},
    Jamie L. Adams\textsuperscript{\rm 2},
    Ruth B. Schneider\textsuperscript{\rm 2},
    E. Ray Dorsey\textsuperscript{\rm 2},
    Ehsan Hoque\textsuperscript{\rm 1}\thanks{Corresponding author. \textbf{Email}: mehoque@cs.rochester.edu}
}
\affiliations {
    \textsuperscript{\rm 1} University of Rochester, United States\\
    \textsuperscript{\rm 2} University of Rochester Medical Center, United States
}

\usepackage{bibentry}
\begin{document}
\pagestyle{plain}
\maketitle

\begin{abstract}
We present an artificial intelligence (AI) system to remotely assess the motor performance of individuals with Parkinson's disease (PD). In our study, 250 global participants performed a standardized motor task involving finger-tapping in front of a webcam. To establish the severity of Parkinsonian symptoms based on the finger-tapping task, three expert neurologists independently rated the recorded videos on a scale of 0 to 4, following the Movement Disorder Society Unified Parkinson's Disease Rating Scale (MDS-UPDRS). The inter-rater reliability was excellent, with an intra-class correlation coefficient (ICC) of 0.88. We developed computer algorithms to obtain objective measurements that align with the MDS-UPDRS guideline and are strongly correlated with the neurologists' ratings. Our machine learning model trained on these measures outperformed two MDS-UPDRS certified raters, with a mean absolute error (MAE) of 0.58 points compared to the raters' average MAE of 0.83 points. However, the model performed slightly worse than the expert neurologists (0.53 MAE). The methodology can be replicated for similar motor tasks, providing the possibility of evaluating individuals with PD and other movement disorders remotely, objectively, and in areas with limited access to neurological care. 
\end{abstract}

\section{Introduction}

Parkinson's disease (PD) is the fastest-growing neurological disease, and currently, it has no cure. Regular clinical assessments and medication adjustments can help manage the symptoms and improve the quality of life. Unfortunately, access to neurological care is limited, and many individuals with PD do not receive proper treatment or diagnosis. For example, in the United States, an estimated 40\% of individuals aged 65 or older living with PD do not receive care from a neurologist~\cite{willis2011neurologist}. Access to care is much scarce in developing and underdeveloped regions, where there may be only one neurologist per millions of people \cite{kissani2022does}. Even for those with access to care, arranging clinical visits can be challenging, especially for older individuals living in rural areas with cognitive and driving impairments.

The finger-tapping task\footnote{The finger-tapping task requires an individual to repeatedly tap their thumb finger with their index finger as fast and as big as possible.} is commonly used in neurological exams to evaluate bradykinesia (i.e., slowing of movement) in upper extremities, which is a key symptom of PD \cite{hughes1992accuracy}. Videos of finger-tapping tasks have been used to analyze movement disorders like PD in prior research. However, the videos are often collected from a few participants ($<20$) \cite{khan2014computer}, or the studies only provide binary classification (e.g., slight vs. severe Parkinsonian symptoms; Parkinsonism vs. non-Parkinsonism) and do not measure PD severity \cite{williams2020supervised, nunes2022automatic}. Additionally, existing models lack interpretability, making it difficult to use them in clinical settings. Most importantly, the videos are noise-free as they are recorded in a clinical setting with the guidance of experts. Machine learning models trained on clean data may not perform effectively if the task is recorded in a noisy home environment due to the models' susceptibility to data shift~\cite{quinonero2008dataset}. Consequently, these models may not enhance access to care for Parkinson's disease.

\begin{figure*}
\centering

\includegraphics[width=0.80\linewidth]{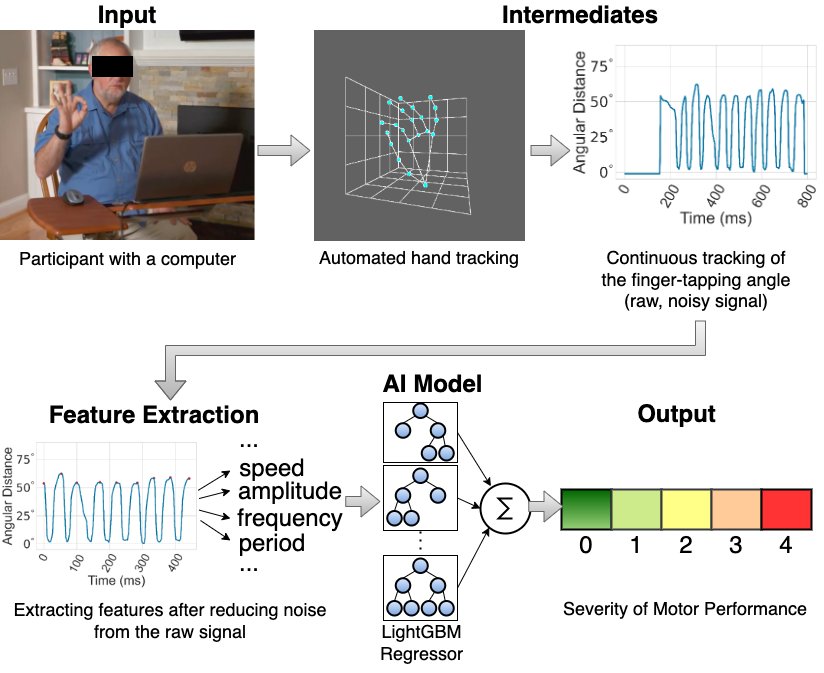}
%\caption{Instead of using complex models that try to fuse multiple modalities, we rely on a single pre-trained language model. As input to language model, we extend the utterance text with visual-text and acoustic-text extracted from communication behavior videos. We group the nonverbal (visual and acoustic) features that frequently appear together into clusters, and use textual descriptions of the clusters to extract these nonverbal-texts.\commentG{rearrange the sentences}}
\caption{\textbf{Overview of the AI-based system for assessing the severity of motor performance.} Anyone can perform the finger-tapping task in front of a computer webcam. The system employs a hand-tracking model to locate the key points of the hand, enabling a continuous tracking of the finger-tapping angle incident by the thumb finger-tip, the wrist, and the index finger-tip. After reducing noise from the time-series data of this angle, the system computes several objective features associated with motor function severity. The AI-based model then utilizes these features to assess the severity score automatically.}
\label{fig:overview}
\end{figure*}

Imagine anyone from anywhere in the world could perform a motor task (i.e., finger-tapping) using a computer webcam and get an automated assessment of their motor performance severity. This presents several challenges: collecting a large amount of data in the home environment, developing interpretable computational features that can be used as digital biomarkers to track the severity of motor functions, and developing a platform where (elderly) people can complete the tasks without direct supervision. In this paper, we address these challenges by leveraging AI-driven techniques to derive interpretable metrics related to motor performance severity and apply them across 250 global participants performing the task mostly from home. Three experts\footnote{All the experts are US neurologists with at least five years of experience in PD clinical studies and actively consult PD patients.} and two non-experts\footnote{The non-experts are MDS-UPDRS certified raters but do not actively consult PD patients. One of the non-experts holds a medical degree from a non-US institution and has actively engaged in multiple PD clinical studies. The other non-expert is a second-year neurology resident who has been active in movement disorder research for 10 years.} rated the severity of motor performance watching these videos, using the Movement Disorder Society Unified Parkinson’s Disease Rating Scale (MDS-UPDRS). Our proposed interpretable, clinically relevant features highly correlate with the experts’ ratings. An AI-based model was trained on these features to assess the severity score automatically, and we compared its performance against both expert and non-expert clinicians. Figure \ref{fig:overview} presents an illustrative overview of our system.

\begin{figure*}
\centering

\includegraphics[width=\linewidth]{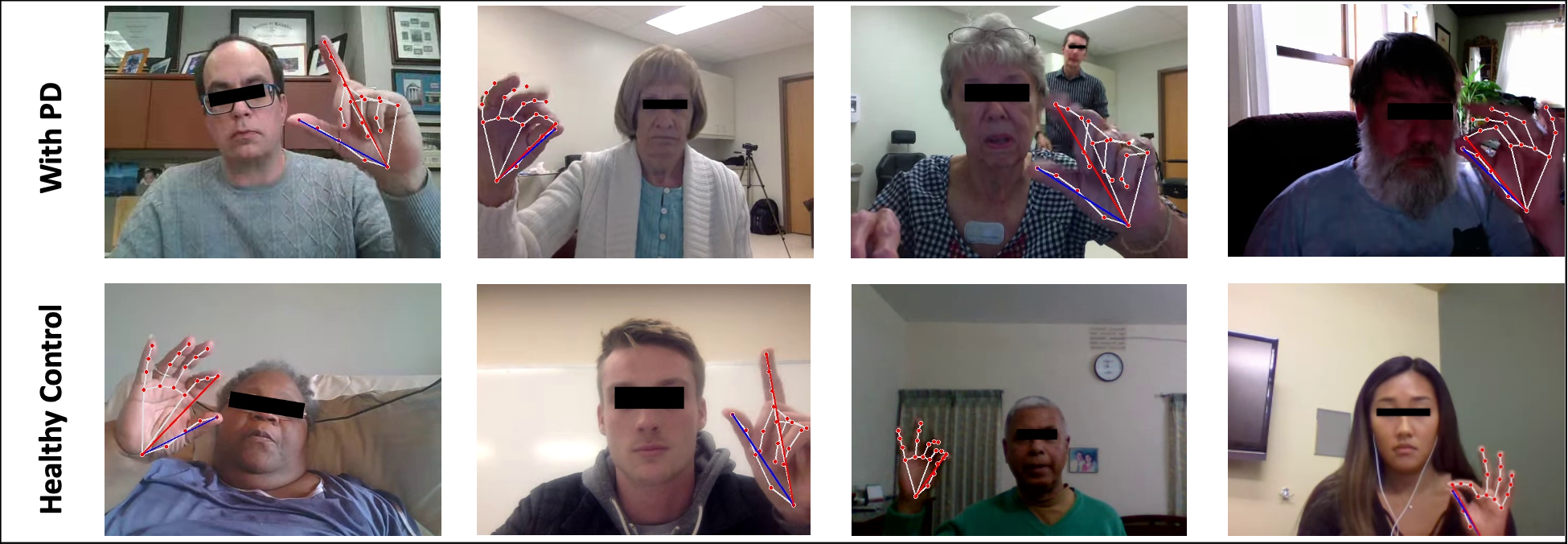}
%\caption{Instead of using complex models that try to fuse multiple modalities, we rely on a single pre-trained language model. As input to language model, we extend the utterance text with visual-text and acoustic-text extracted from communication behavior videos. We group the nonverbal (visual and acoustic) features that frequently appear together into clusters, and use textual descriptions of the clusters to extract these nonverbal-texts.\commentG{rearrange the sentences}}
\caption{\textbf{Data collection.} The participants, both those with Parkinson's disease (PD) and healthy controls, performed the task primarily in a noisy home environment without any clinical supervision. The dataset includes blurry videos caused by poor internet connection, videos where participants had difficulty following instructions, and videos with overexposed or underexposed backgrounds. These issues are common when collecting data from home, particularly from an aged population that may be less familiar with technology than other age groups.}
\label{fig:examples}
\end{figure*}

\section{Results}
%Note: sub-headings will be removed. Just keeping it to assist the writing process.
\subsection*{Data}

We obtained data from 250 global participants (172 with PD, 78 control) who completed a finger-tapping task with both hands (see Figure \ref{fig:examples} for examples). Participants used a web-based tool \cite{langevin2019park} to record themselves with a webcam primarily from their homes. Demographic information for the participants is presented in Table \ref{tab:demography}. %The dataset is fairly balanced in terms of gender and includes participants from a broad age range (18-86 years). However, we had limited participation from non-majority races -- which we aim to improve in the future.

\begin{table}[!htb]
\centering
\resizebox{1.0\columnwidth}{!}{%
\begin{tabular}{|lllll|}
\hline
\multicolumn{2}{|l}{\textbf{Characteristics}}                                                 & \textbf{With PD}      & \textbf{Without PD}  & \textbf{Total}                 \\ \hline
\multicolumn{2}{|l}{\begin{tabular}[c]{@{}l@{}}Number of \\ Participants, \\ n \end{tabular} }                                  & 172 & 78 & \textbf{250}  \\ \hline
\multicolumn{5}{|l|}{Sex, n (\%)}                                                                                                      \\
 & Male                                                                              & 109 (63.4\%) & 28 (35.9\%) & \textbf{137 (54.8\%)} \\
 & Female                                                                            & 63 (36.6\%)  & 50 (64.1\%) & \textbf{113 (45.2\%)} \\ \hline
\multicolumn{5}{|l|}{Age in years (range: 18 - 86, mean: 62.13), n (\%)}                                                                                 \\
 & \textless{}20                                                                     & 0 (0.0\%)    & 3 (3.8\%)   & \textbf{3 (1.2\%)}    \\
 & 20-29                                                                             & 0 (0.0\%)    & 10 (12.8\%)  & \textbf{10 (4.0\%)}   \\
 & 30-39                                                                             & 1 (0.6\%)    & 3 (3.8\%)   & \textbf{4 (1.6\%)}    \\
 & 40-49                                                                             & 5 (2.9\%)    & 6 (7.7\%)   & \textbf{11 (4.4\%)}   \\
 & 50-59                                                                             & 34 (19.8\%)  & 14 (18.0\%)  & \textbf{48 (19.2\%)}  \\
 & 60-69                                                                             & 64 (37.2\%)  & 30 (38.5\%) & \textbf{94 (37.6\%)}  \\
 & 70-79                                                                             & 62 (36.0\%)  & 12 (15.4\%)  & \textbf{74 (29.6\%)}  \\
 & \textgreater{}=80                                                                 & 6 (3.5\%)    & 0 (0.0\%)   & \textbf{6 (2.4\%)}    \\ \hline
\multicolumn{5}{|l|}{Race, n (\%)}                                                                                                        \\
 & White                                                                             & 161 (93.6\%) & 69 (88.5\%) & \textbf{230 (92.0\%)} \\
 & \textcolor{blue}{Asian}                                                                             & 2 (1.2\%)    & 5 (6.4\%)   & \textbf{7 (2.8\%)}    \\
 & \begin{tabular}[c]{@{}l@{}}Black or \\ African \\ American\end{tabular}           & 1 (0.6\%)    & 2 (2.6\%)   & \textbf{3 (1.2\%)}    \\
 & \begin{tabular}[c]{@{}l@{}}\textcolor{blue}{American} \\ \textcolor{blue}{Indian or} \\ \textcolor{blue}{Alaska} \\ \textcolor{blue}{Native}\end{tabular} & 2 (1.2\%)    & 0 (0.0\%)   & \textbf{2 (0.8\%)}    \\
 & Others                                                                            & 1 (0.6\%)    & 0 (0.0\%)   & \textbf{1 (0.4\%)}    \\
 & \textcolor{blue}{No mention}                                                                   & 5 (2.9\%)    & 2 (2.6\%)   & \textbf{7 (2.8\%)}    \\ \hline
 \multicolumn{5}{|l|}{Recording Environment, n (\%)}                                                                                                      \\
 & Home                                                                              & 140 (81.4\%) & 59 (75.6\%) & \textbf{199 (79.6\%)} \\
 & Clinic                                                                            & 25 (14.5\%)  & 17 (21.8\%) & \textbf{42 (16.8\%)} \\
 & Unknown                                                                            & 7 (4.1\%)  & 2 (2.6\%) & \textbf{9 (3.6\%)} \\\hline
\end{tabular}%
}
\caption{\textbf{Demographic characteristics of the participants.} %The dataset is fairly balanced in terms of sex and includes participants from a broad age range (18-86 years). However, we had limited participation from non-majority races -- which we aim to improve in the future.
}
\label{tab:demography}
%\vspace{-3mm}
\end{table}

Following the MDS-UPDRS guidelines, we considered each participant's left and right-hand finger-tapping as two separate videos. All these $250 \times 2 = 500$ videos are rated by three expert neurologists with extensive experience providing care to individuals with PD and leading PD research studies. However, after undertaking manual and automated quality assessments, we removed 11 videos from the dataset. Ultimately, we had 489 videos for analysis (244 videos for the left hand and 245 for the right hand). We obtained the ground truth severity score a) by majority agreement when at least two experts agreed on their ratings (451 cases), or b) by taking the average of three ratings and rounding it to the nearest integer when no majority agreement was found (38 cases). 

\subsection*{Rater Agreement}
\begin{figure*}
\centering

\includegraphics[width=\linewidth]{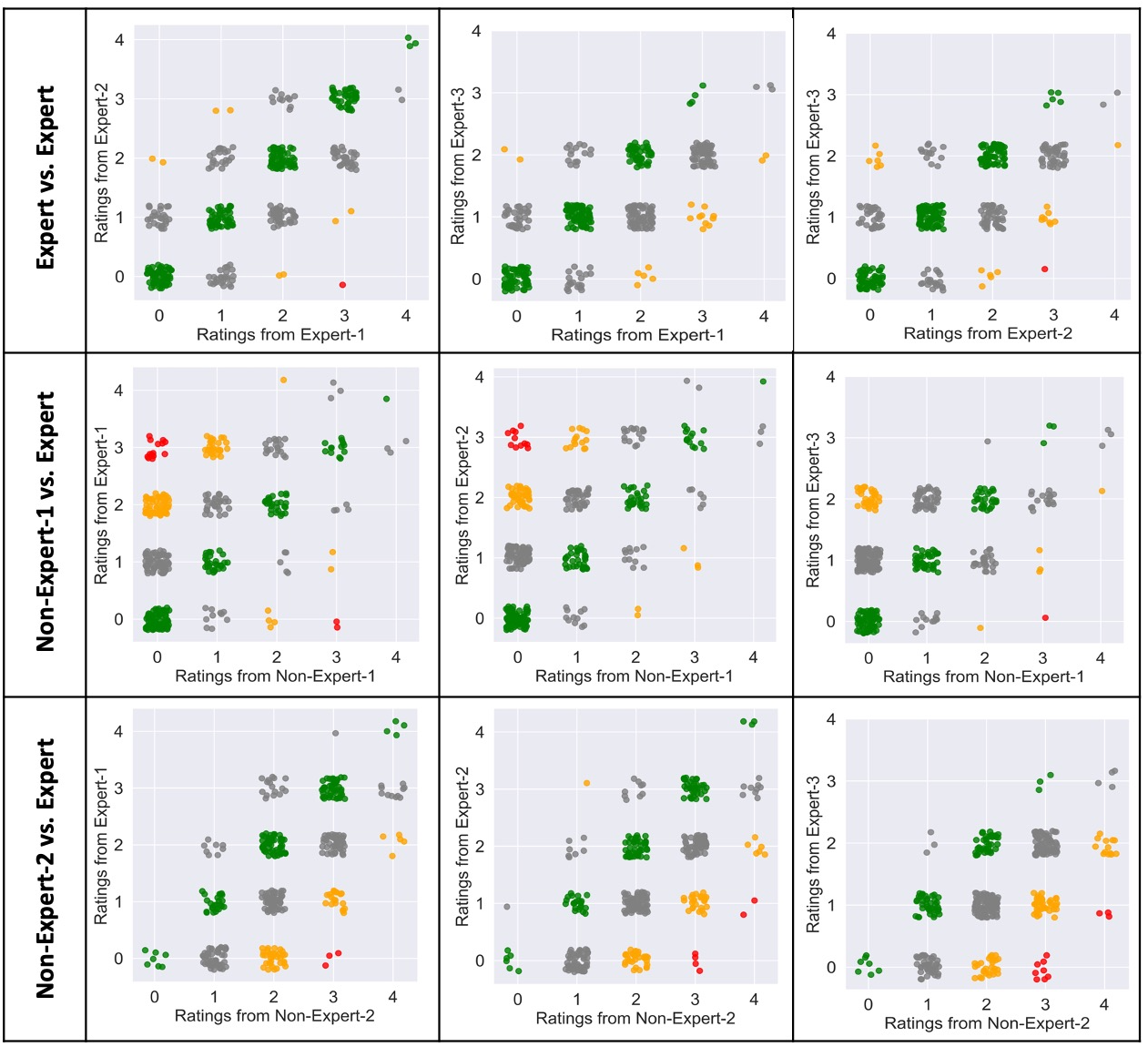}
%\caption{Instead of using complex models that try to fuse multiple modalities, we rely on a single pre-trained language model. As input to language model, we extend the utterance text with visual-text and acoustic-text extracted from communication behavior videos. We group the nonverbal (visual and acoustic) features that frequently appear together into clusters, and use textual descriptions of the clusters to extract these nonverbal-texts.\commentG{rearrange the sentences}}
\caption{\textbf{An overview of how the experts and the non-experts agreed on their ratings.} Green dots indicate two raters having a perfect agreement, while grey, orange, and red dots imply a difference of 1, 2, and 3 points, respectively. We did not observe any 4 points rating difference. The high density of green and gray dots and an ICC score of 0.88 verifies that the experts demonstrated high inter-rater agreement among themselves, and the finger-tapping task can be reliably rated when recorded from home. However, the non-experts were less reliable than the experts, demonstrating moderate agreement with the three expert raters (the average ICC of a non-expert’s ratings and the ratings from the three experts were 0.72, 0.74, and 0.70, respectively.)}
\label{fig:agreement}
\end{figure*}

The three expert neurologists demonstrated good agreement on their ratings, as measured by a) Krippendorff's alpha score of 0.69 and b) Intra-class correlation coefficient (ICC) score of 0.88 (95\% confidence interval: [0.86, 0.90]).  %Krippendorff's alpha was reported instead of Cohen Kappa score, since the severity ratings were ordinal in nature (0: normal, 1: slight, 2: mild, 3: moderate, 4: severe). 
Figure \ref{fig:agreement} provides an overview of pair-wise agreement between expert raters.
All three experts agreed in 30.7\% of the videos, and at least two agreed in 93\% videos. The three raters showed a difference of no more than 1 point from the ground truth in 99.2\%, 99.5\%, and 98.2\% of the cases, respectively. These metrics suggest that the experts can reliably rate our videos recorded from home environments. 

\subsection*{Features as Digital Biomarkers}
We quantified 47 features measuring several aspects of the finger-tapping task, including speed, amplitude, hesitations, slowing, and rhythm. We also quantified how much an individual's wrist moves using 18 features. For each feature, Pearson's correlation coefficient ($r$) is measured to see how the feature is correlated with the ground truth severity score, along with a statistical significance test (significance level $\alpha = 0.01$). We found that 22 features were significantly correlated with the severity scores, which reflects their promise for use as digital biomarkers of symptom progression. Table \ref{tab:features} shows the top 10 features with the highest correlation. These features are clinically meaningful as they capture several aspects of speed, amplitude, and rhythm (i.e., regularity) of the finger-tapping task, which are focused on the MDS-UPDRS guideline for scoring PD severity.

\begin{table}[ht]
\centering
\resizebox{1.05\columnwidth}{!}{%
\begin{tabular}{|lllll|}
\hline
\textbf{Feature}        & \textbf{Statistic}                                                                             & \textbf{r}     & \textbf{p-value}    & \textbf{Rank} \\ \hline
\multirow{3}{*}{speed}     & inter-quartile range                                                                   & -0.56 & $10^{-33}$ & 1    \\
                           & median                                                                                 & -0.52 & $10^{-27}$ & 2    \\
                           & maximum                                                                                & -0.32 & $10^{-10}$ & 6    \\ \hline
\multirow{2}{*}{amplitude} & median                                                                                 & -0.50 & $10^{-25}$ & 3    \\
                           & maximum                                                                                & -0.41 & $10^{-17}$ & 4    \\ \hline
\multirow{2}{*}{frequency} & inter-quartile range                                                                   & 0.32  & $10^{-10}$ & 5    \\
                           & standard deviation                                                                     & 0.29  & $10^{-8}$  & 8    \\ \hline
\multirow{3}{*}{period}    & entropy (i.e., irregularity)                                                           & 0.32  & $10^{-10}$ & 7    \\
                           & \begin{tabular}[c]{@{}l@{}}variance (normalized \\ by the average period)\end{tabular} & 0.28  & $10^{-8}$  & 9    \\
                           & inter-quartile range                                                                   & 0.27  & $10^{-7}$  & 10   \\ \hline
\end{tabular}%
}
\caption{\textbf{Features most correlated with the ground truth severity scores for the finger-tapping task.} r and p-value indicate Pearson's correlation coefficient and significance level, respectively. %These features are clinically meaningful as they capture several aspects of speed, amplitude, and rhythm (i.e., regularity) of the finger-tapping task which are focused on the MDS-UPDRS guideline for scoring PD severity.
}
\label{tab:features}
\end{table}

Traditionally, human evaluators cannot constantly measure the finger-tapping speed. Instead, they count the number of taps the participant has completed within a specific time (e.g., three taps per second). However, in our case, the videos were collected at 30 frames per second rate, thus allowing us to track the fingertips 30 times per second and develop a continuous measure of speed. The former approach, the number of finger taps completed in unit time, is termed as ``frequency'' throughout the paper, and ``speed'' (and ``acceleration'') refers to the continuous measure (i.e., movement per frame). Similarly, ``period'' refers to the time it takes to complete a tap, and thus, is a discrete measure. In addition, finger tapping amplitude is measured by the maximum distance between the thumb and index-finger tips during each tap. Since linear distance can vary depending on how far the participant is sitting from the camera, we approximated amplitude using the maximum angle incident by three key points: the thumb-tip, the wrist, and the index fingertip. As we see in Table \ref{tab:features}, several statistical measures of continuous speed are significantly correlated with PD severity. These granular computations are only attainable using automated video analysis, which, to our knowledge, was missing in prior literature.

%The most significant feature is the inter-quartile range (IQR) of finger tapping speed. It measures one's ability to demonstrate a range of speeds (measured in a continuous way) while performing finger-tapping and is negatively correlated with PD severity. As the index finger is about to touch the thumb finger, an individual needs to decelerate and operate at a low speed. Again, when the index finger moves away from the thumb finger after tapping, one needs to accelerate and operate at a high speed. A higher range implies a higher difference between the maximum and minimum speed, which can be thought of as someone having more control over the variation of speed, and thus, is linked to healthier motor functions. Additionally, the median and the maximum finger-tapping speed, as well as the median and the maximum finger-tapping amplitudes have strong negative correlations with PD severity. Finally, IQR and standard deviation of finger tapping frequency, as well as entropy, variance, and IQR of tapping periods are found to have strong positive correlations with PD severity, as they all indicate the absence of regularity in the tapping amplitude and periods. These findings align with prior clinical studies reporting that individuals with Parkinson’s have a slower and less rhythmic finger tapping compared to those without the condition \cite{kim2011quantification}.

% \usepackage{multirow}

%Also, show how the top-5 features vary across different gender, age, and pd vs non-pd groups.

%\vspace{-4mm}

\subsection*{Performance of Non-Expert Clinicians}
%When an individual does not have access to a neurologist, talking to a primary care physician (PCP), or a clinician might be their only options. For these reasons it is critical to assess how a clinician with limited expertise in movement disorders or parkinson's disease may perform compared to the experts. We recruit a MDS-UPDRS certified clinician who had a foreign MBBS (Bachelor of Medicine, Bachelor of Surgery) degree, had some expertise rating the severity of PD symptoms in a clinical setting, but not a movement disorder specialist. The non-expert clinician was asked to rate the same videos rated by three experts, and we observe moderate reliability of her ratings, as the intra-class correlation (ICC) coefficient of her ratings and the ground truth ratings was 0.70. The the non-expert deviated from the ground truth severity score by 0.79 points on average, and the Pearson's correlation coefficient between the non-expert ratings and the ground truth severity scores was 0.5456. This substantial difference between the reliability of expert and non-expert ratings depicts the difficulty of rating PD severity, and calls for action items to improve the access and equity of neurological healthcare.

\begin{figure*}
  \begin{subfigure}[t]{.45\textwidth}
    \centering
    \includegraphics[width=0.8\linewidth]{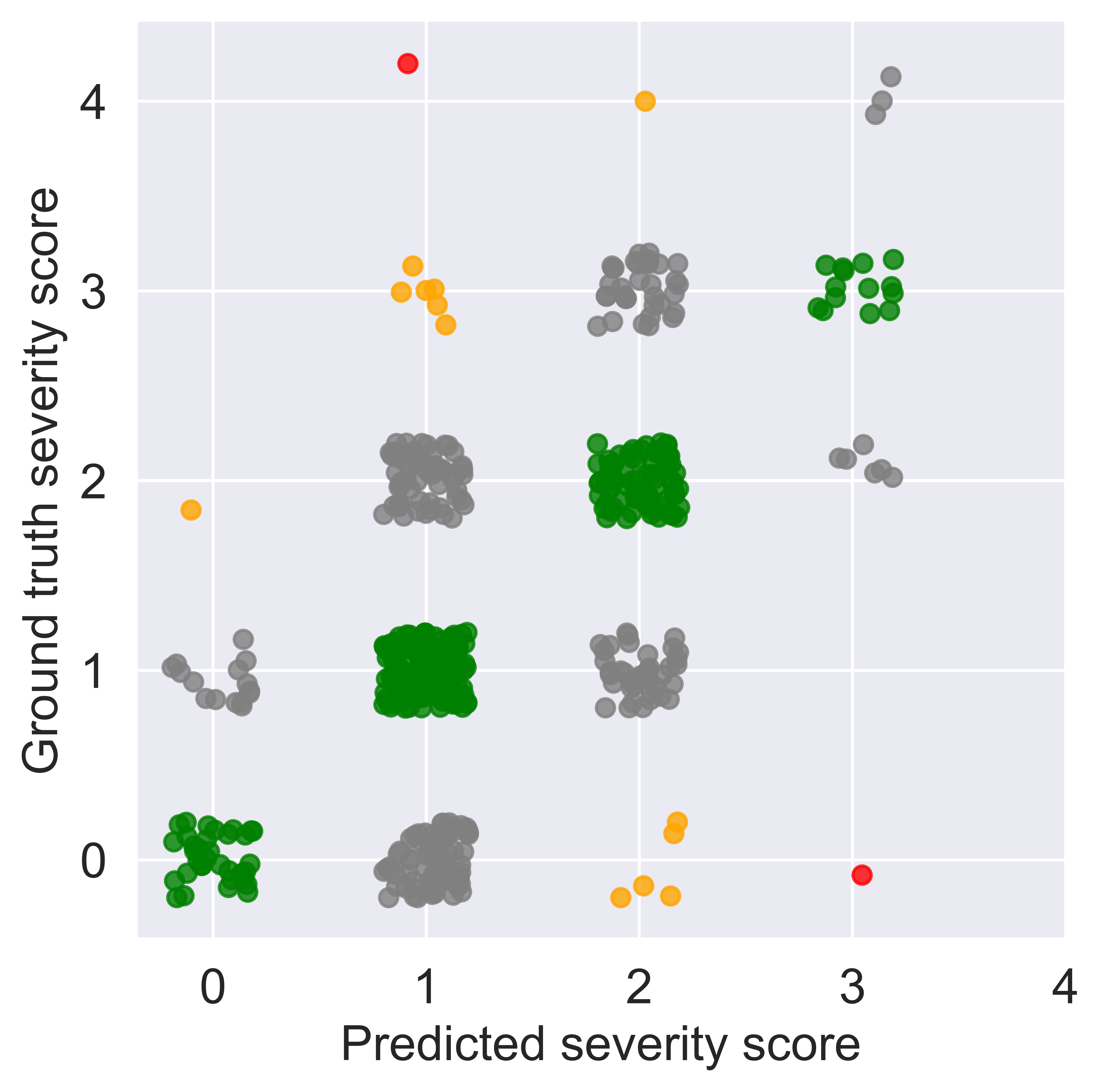}
    \caption{}
    %\caption{Agreement between the predicted severity and the ground truth scores.}
    \label{fig:per1}
  \end{subfigure}
  \hfill
  \begin{subfigure}[t]{.45\textwidth}
    \centering
    \includegraphics[width=0.88\linewidth]{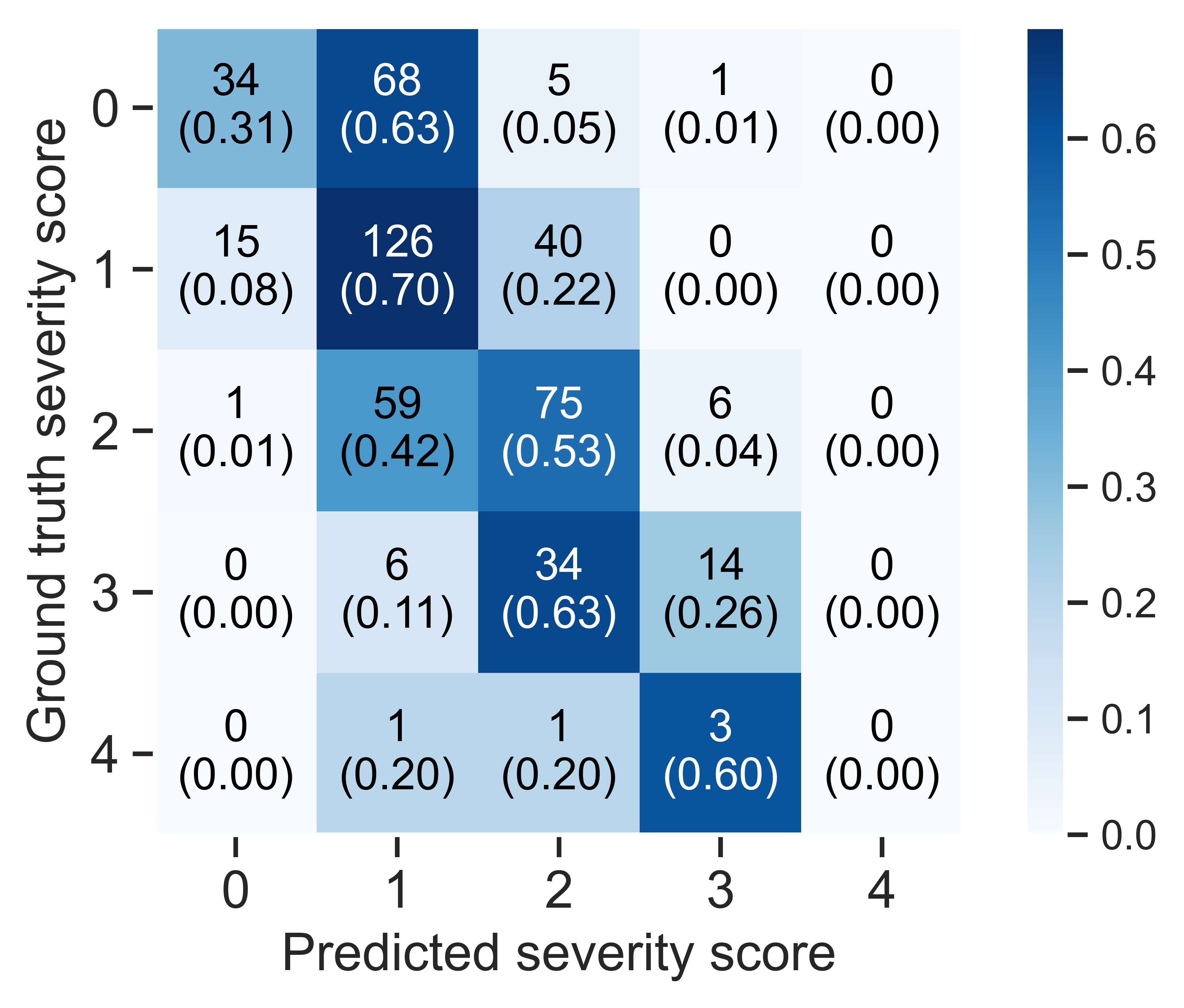}
    \caption{}
    %\caption{Confusion matrix for model predictions.}
    \label{fig:confusion}
  \end{subfigure}

  \bigskip

  \begin{subfigure}[t]{.45\textwidth}
    \centering
    \includegraphics[width=\linewidth]{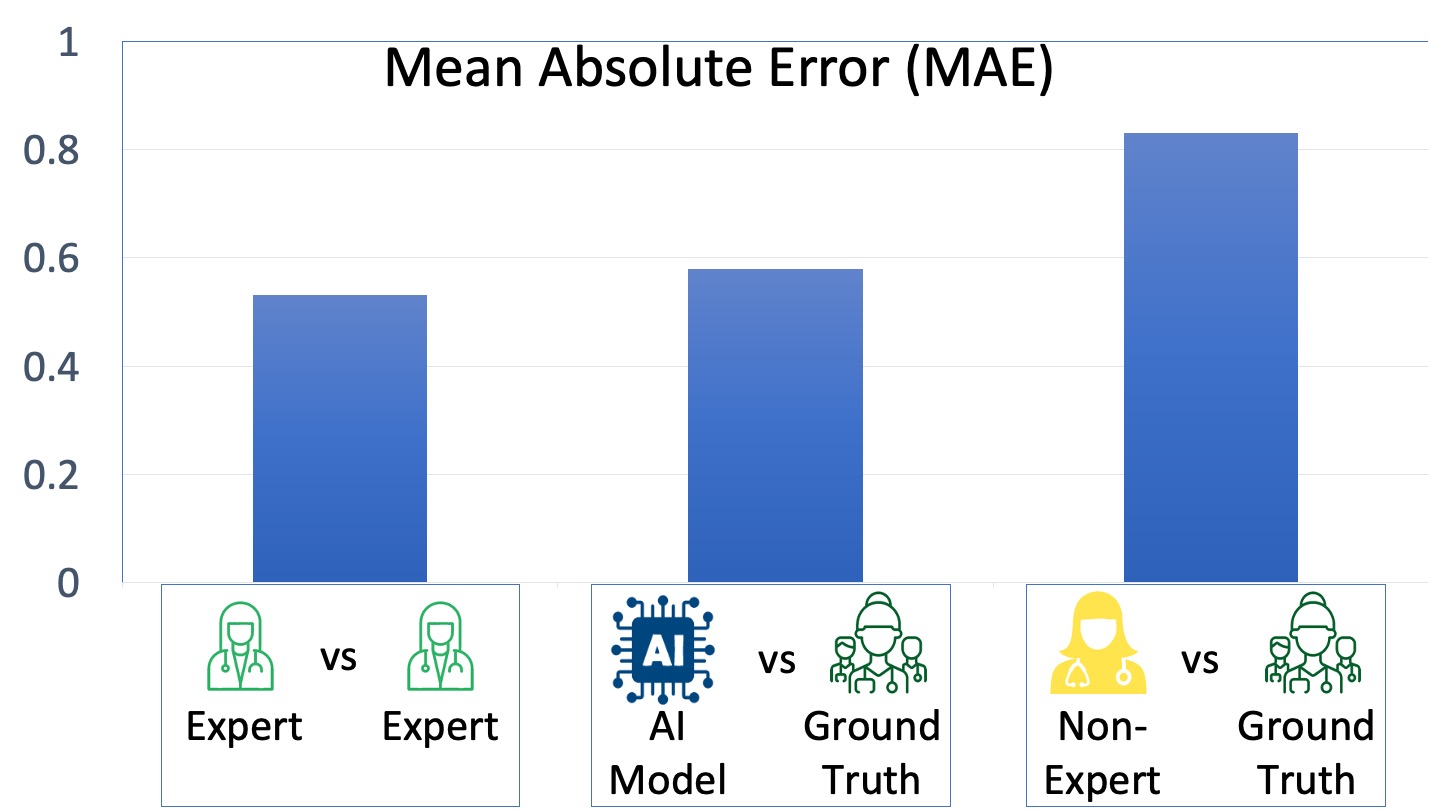}
    %\caption{Comparison of error.}
    \caption{}
    \label{fig:mae}
  \end{subfigure}
  \hfill
  \begin{subfigure}[t]{.45\textwidth}
    \centering
    \includegraphics[width=\linewidth]{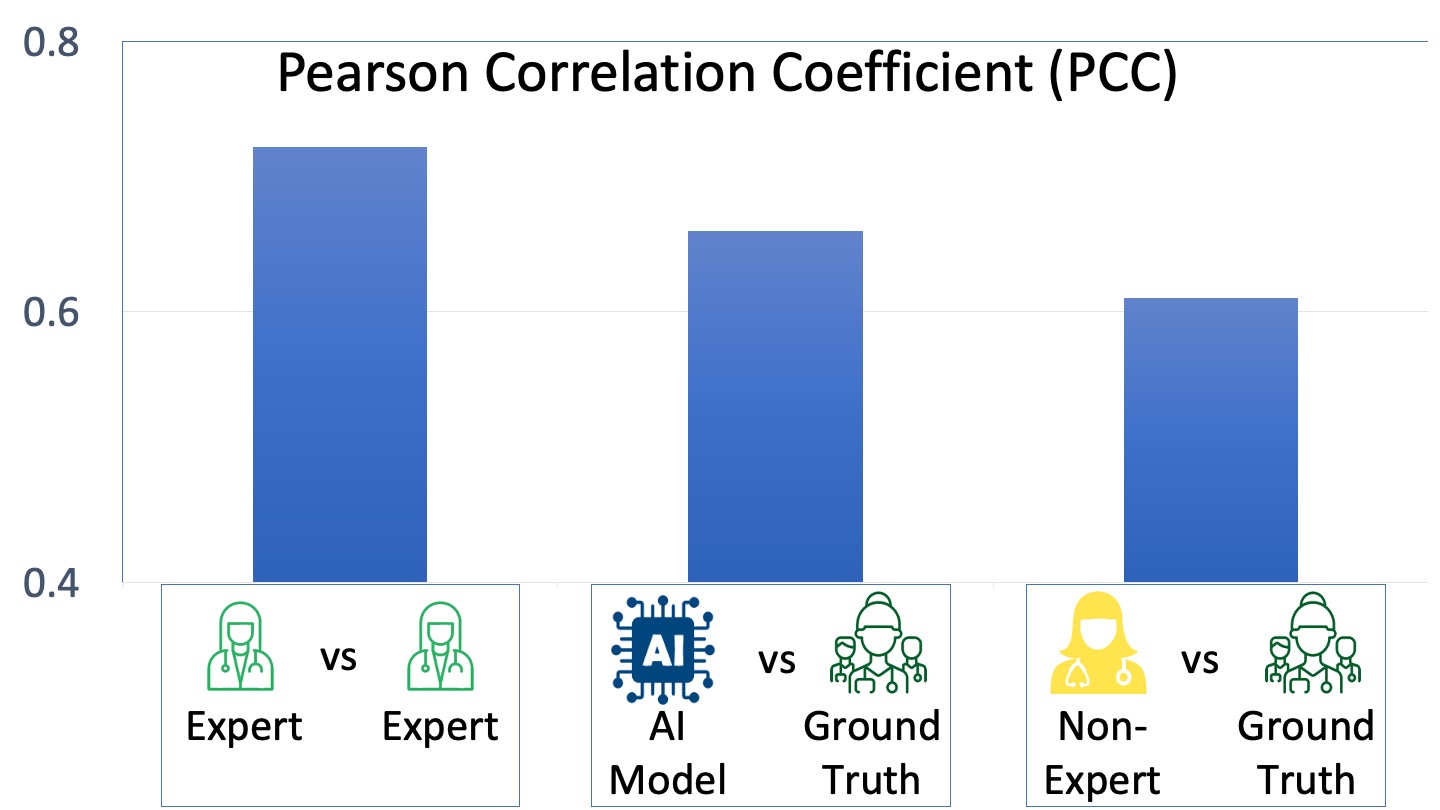}
    %\caption{Comparison of correlation.}
    \caption{}
    \label{fig:mae}
  \end{subfigure}

  \caption{\textbf{Model performance.} (a) We observe good agreement between the predicted severity and the ground truth scores. Green dots indicate correct predictions, while grey, orange, and red dots imply a difference of 1, 2, and 3 points between the predicted and actual scores. We did not observe any 4 points rating difference. (b) The confusion matrix presents the agreement numerically. (c) The mean absolute error (MAE) measures the difference between two ratings. The model incurs slightly higher MAE than an average expert but substantially lower MAE than the non-experts. (d) Pearson correlation coefficient (PCC) measures the correlation between two sets of ratings. The model's predicted severity ratings are more correlated with the ground truth scores than the non-experts' (higher PCC) but less correlated than an average expert's (lower PCC) ratings.}
  \label{fig:performance}
\end{figure*}

When an individual lacks access to a neurologist with expertise in movement disorders, they may consult with a non-specialist clinician. Thus, it is critical to assess how a clinician with limited expertise in movement disorders or Parkinson's disease may perform compared to experts in this field. To this end, we recruited two investigators. The first investigator (referred as non-expert-1 throughout the manuscript) had completed an MBBS \footnote{MBBS: Bachelor of Medicine, Bachelor of Surgery} but not additional medical training (e.g., residency), was certified to administer the MDS-UPDRS, and had the experience of rating the severity of PD symptoms in multiple research studies. The second investigator (referred as non-expert-2) is a second-year neurology resident at a reputed medical center in the United States. This investigator has been involved in movement disorders research and clinical trials for 10 years, and takes care of less than 20 PD patients a year. %a clinician who was certified to administer the MDS-UPDRS, had completed an MBBS (Bachelor of Medicine, Bachelor of Surgery) degree outside the U.S., and had experience rating the severity of PD symptoms in multiple clinical studies. However, the clinician did not attend a neurology residency, and does not provide consultation to patients with movement disorders. 
We asked the non-expert clinicians to rate the same videos that the three experts had rated. We observed moderate reliability in their ratings, with an average intra-class correlation coefficient (ICC) of 0.75 compared to the ground truth scores (Figure \ref{fig:agreement}). On average, the non-experts' ratings deviated from the ground truth severity score by 0.83 points, and the average Pearson's correlation coefficient (PCC) between the non-experts' ratings and the ground truth severity scores was 0.61.

\subsection*{Modeling the Severity Rating}

We employed a machine learning model (i.e., LightGBM regressor \cite{ke2017lightgbm}) that predicts the severity of PD symptoms based on the extracted features from a finger-tapping video. To evaluate the model, we implemented a leave-one-patient-out cross validation approach, that leaves all data (e.g., videos of both left and right hand) associated with a particular patient as a test set and the model is trained with the remaining data. The evaluation uses multiple iterations to ensure that the performance is validated once for each patient. %To evaluate the model, we implemented a stratified $k$-fold cross-validation technique with $k=10$. This involved dividing the dataset into 10 equally sized folds, with each fold containing examples from all severity classes in the same ratio as the entire dataset. 9 out of 10 folds were used to train the XGBoost model, and the model was evaluated on the remaining fold. The average performance of 20 random iterations was reported. Stratified cross-validation ensures that each fold of the cross-validation process has a representative proportion of examples from each class in the dataset, and the reported evaluation metrics accurately reflect the model's performance on the entire dataset. This is particularly important in imbalanced datasets where the number of examples in each class can vary widely.
The severity rating predicted by the model is a continuous value, ranging from 0 to 4. We employed several standard metrics used to assess the performance of machine learning models in regression tasks: mean absolute error (MAE), mean squared error (MSE), Kendall rank correlation coefficient (Kendall’s Tau), mean average percentage error (MAPE), Pearson’s correlation coefficient (PCC), and Spearman’s rank correlation coefficient (Spearman’s $\rho$). However, in this manuscript, we primarily focus on MAE and PCC that are the most popular in assessing a model's regression capability. Also, to measure classification accuracy, the continuous prediction is converted to five severity classes (0, 1, 2, 3, and 4) by rounding it to the closest integer. The classification result is reported in Figure \ref{fig:performance} (a, b), showing that the model's predictions largely agree with the ground truth severity scores. The model's reliability in rating the videos is moderate, as indicated by an ICC score of 0.76 (95\% C.I.: [0.71, 0.80]). On average, the model predictions deviated from the ground truth severity scores by 0.58 points, and the Pearson's correlation coefficient between the predictions and the ground truth severity scores was 0.66. Since the ground truth scores were derived from the three experts' ratings, it is natural to find an excellent correlation (PCC = 0.86) and a minimal difference (MAE = 0.27) between an average expert and the ground truth. However, this is an unfair baseline to compare the performance of the model and the non-expert. Instead, we looked at how the expert neurologists agreed with each other to establish a human-level performance of the rating task. On average, any pair of experts differed by 0.53 points from each other's ratings, and their ratings were correlated with PCC = 0.72. In most of the metrics we tested, the LightGBM regression model outperformed the non-expert clinicians but \emph{was outperformed} by the experts (Figure \ref{fig:performance} (c, d)). Please see supplementary information for details.

%\vspace{-6mm} 

% \subsection*{Explanation of Model Predictions}

% \begin{figure}
%     \centering
%     \includegraphics[width=\linewidth]{LaTeX/figures/pd_severity_shap_summary.png}
%     \caption{SHAP scores for salient features. \textcolor{red}{Figure needs improvement. Feature names should align with the ones in the table. We may also consider removing this and just mentioning some of the features the model focused more on.}}
%     \label{fig:PD_severity_shap_summary}
% \end{figure}

% -- \textcolor{red}{WASIF}
\subsection*{Interpretability of Model Predictions}
We used SHapley Additive exPlanations (SHAP) to interpret the outputs of the machine learning model. SHAP values provide a way to attribute a prediction to different features and quantify the impact of each feature on the model output~\cite{lundberg2020local2global}. This information can be useful for understanding how a model makes decisions and for identifying which features are most important for making accurate predictions. We found that important features identified by SHAP align well with our previously identified significant features (see Table \ref{tab:features}). Specifically, SHAP identified the inter-quartile range (IQR) of finger-tapping speed as the most important feature driving the model's predictions, which is the most correlated feature with the ground-truth severity score. Top-10 most important features include finger-tapping speed (IQR), freezing (maximum duration and number of freezing), absence of periodicity, period (variance, minimum), wrist-movement (minimum, median), and frequency (IQR), which are all significantly correlated with the ground-truth severity score (at a significance level, $\alpha = 0.01$). The only top-10 feature that does not correlate significantly with the severity score is the median period (average time taken to complete each tap), which is also underscored in the MDS-UPDRS guideline. These results indicate that the model is looking at the right features while deciding the finger-tapping severity scores from the recorded videos, further suggesting the model's reliability.

\subsection*{Analyzing Bias}
To better evaluate the performance of our model across various demographic groups, we conducted a group-wise error analysis that takes into account factors such as gender, age, and Parkinson's disease diagnosis status. This approach allows us to assess any potential biases or inaccuracies in our model's predictions and make necessary improvements to ensure equitable and accurate results for all users. We combined the model predictions of the samples for each patient when they were in the test set. Additionally, we tracked the demographic attributes associated with each sample, allowing us to evaluate the performance of our model across various demographic groups. 

Our model achieved a mean absolute error (MAE) of 0.60 (std\footnote{std: standard deviation} = 0.48) for male subjects ($n = 267$) and 0.55 (std = 0.39) for female subjects ($n = 222$), indicating relatively accurate predictions for both genders. Furthermore, we conducted statistical test (i.e., two-sample two-tailed t-test) to compare the errors across the two groups and found no significant difference (p-value = 0.21). 

We also performed a similar error analysis on our model's predictions for subjects with PD ($n = 333$) and those without PD ($n = 156$). The model had an MAE of 0.57 for subjects with PD and 0.59 for those without PD. However, we found no significant difference in the errors between these two groups (p-value = 0.61). These results suggest that our model does not exhibit detectable gender bias and performs similarly for PD and non-PD groups. 

Age had a slight negative correlation with the error of model predictions (Pearson's correlation coefficient, $r = -0.06$). However, the correlation was not statistically significant at $\alpha = 0.05$ significance level ($p$-value = $0.20$). This suggests that the performance of the model is not significantly different across the younger and older populations. %These analyses underscore the strength of training models with participants from diverse demography. Although younger participants are less likely to show PD symptoms, it is important that they are still represented in the dataset. Otherwise, the model might end up making more errors for younger individuals. 

%This indicates that, on average, the model tends to perform slightly better for older subjects compared to younger ones. This is understandable, as the dataset contained limited samples from subjects below 50 years. While the correlation coefficient is small, it is still important to take into account the potential impact of age on model performance and consider any necessary adjustments or modifications to ensure equitable performance across all age groups. Future studies may consider adding more younger participants although they are less likely to show PD symptoms.

Finally, we tested whether the model exhibits racial bias. Since we did not have enough representation from all the races, the analysis only focused on white vs. non-white population. Our model had an MAE of 0.57 (std = 0.45) for white subjects ($n = 452$), while the MAE was 0.65 (std = 0.41) for non-white subjects ($n = 37$). Although the difference was not statistically significant (p-value = 0.29) at 95\% confidence level, the model seems to perform slightly worse for the non-white population.

\subsection*{Impact of Video Quality}
Since most of the video recordings are collected from participants' homes, the quality of the videos may vary widely. Several factors, such as the lighting condition of the recording environment, quality of the data capturing devices (i.e., webcams and internet browsers), participants' cognitive ability and understanding of the task, surrounding noise (e.g., multiple persons being visible in the recording frame) can affect the quality of the recorded videos. Hence, it is essential to analyze how video quality influences clinical ratings, the performance of the pose estimation model used in this study, and, eventually, the model performance.

When providing finger-tapping severity ratings for each video, each expert neurologist identified cases where the video was difficult to rate due to quality issues or the participant's inability to follow the task appropriately. We grouped the videos into two categories: high-quality videos, which none of the experts had difficulty rating, and low-quality videos, which at least one expert had difficulty rating.

\textbf{Impact on rater agreement:} We examined whether the agreement between the expert raters varied depending on the quality of the videos. For the high-quality videos ($n = 385$), the measure of inter-rater reliability, known as the intra-class correlation coefficient (ICC), was found to be 0.879 (95\% confidence interval [CI] = [0.86, 0.90]). On the other hand, the ICC for the low-quality videos ($n = 104$) was 0.806 (95\% CI = [0.73, 0.86]). Although this difference is not statistically significant (at 95\% confidence level) due to overlap in the confidence intervals for the two groups, it suggests that the video quality may have slightly impacted the agreement among the expert raters. We also ran a Chi-square test of independence and found no significant association between finding a majority agreement among the experts and the quality of the videos at a 95\% confidence interval (test-statistic = 0.9965, p-value = 0.318, degree of freedom = 1). In addition, there were 38 videos (out of 489) where the experts disagreed by at least 2 points. 27 (71\%) out of these 38 videos were marked as high-quality by all experts, whereas only 11 videos (29\%) were marked as low-quality by one or more experts.

\textbf{Impact on pose-estimation model:} Next, we tried to evaluate whether video quality impacts the performance of the pose estimation model used in this study (i.e., MediaPipe). For each frame in the video, MediaPipe provides a ``hand presence score'' while estimating the coordinates of the hand key-points. The hand presence score is a measure of the confidence of the model's ability to track the hand in the current frame. The score is a number between 0 and 1, where 0 indicates no confidence and 1 indicates high confidence. A high hand presence score indicates that the model is confident in its ability to track the hand. This means that the landmarks are likely to be accurate and that the pose estimation is likely to be correct. A low hand presence score indicates that the model is not confident in its ability to track the hand. Therefore, to estimate the pose estimation performance on a video, we used the average hand presence score across all the frames in the video after dropping the starting and ending frames that do not contain the hand of interest. The mean hand presence scores were 0.967 and 0.962 respectively across the high-quality ($n = 385$) and low-quality videos ($n = 104$). Based on these results, we did not observe a notable difference in the performance of MediaPipe hand tracking between the high-quality and low-quality videos. This finding was supported by statistical analysis using a two-sample two-tailed t-test, which yielded a p-value of 0.36 with test statistic, t = 0.91.

Additionally, we manually injected noise to randomly selected 86 ``high-quality'' (43 videos of right-hand and 43 of left-hand) videos from our dataset to see how different types of noise impacts the confidence (hand presence scores) of MediaPipe. Specifically, we applied different levels of blurring by performing a low-pass filter operation on each frame with two different kernel sizes ($3 \times 3$ (slight blurring) and $9 \times 9$ (substantial blurring), and also injected different levels of Gaussian noise with zero mean and two different standard deviations (25 (low amount of noise), and 40 (high amount of noise)). As a result, each video had five distinct versions: the original video, slightly blurred, substantially blurred, low amount of Gaussian noise, and high amount of Gaussian noise. In general, MediaPipe had lower confidence scores for the videos with injected noise. For example, the mean confidence score for the original videos was 0.96 (std = 0.031) while the mean for the slightly blurred and substantially blurred videos were 0.958 (std = 0.033) and 0.935 (std = 0.08) respectively. Similarly, the mean scores for the videos with a low amount of added noise and high amount of added noise were 0.897 (std = 0.124) and 0.784 (std = 0.170) respectively. The difference in MediaPipe confidence scores was not significant between the original and slightly blurred videos. However, rest of the group-wise differences were statistically significant as validated by paired sample t-tests (please see Supplementary Table 5 for test statistics).

\textbf{Impact on model performance:} Finally, we also evaluate whether the model performs poorly for the low-quality videos. Specifically, we evaluate the prediction errors of the model across two groups of videos (high-quality and low-quality) and run two-sample t-tests to see whether there is any significant difference in prediction errors across these two groups. The mean absolute errors of the model were 0.581 (std = 0.43) and 0.578 (std = 0.506) respectively across the high and low quality videos. Therefore, we observed no significant difference in performance across these two groups (t = 0.064, p-value = 0.95) based on two-sample t-test. Also, we analyzed the cases where the model predictions were off by more than 1.5 points (resulting in at least 2 points difference when the continuous predictions are converted into severity classes). Out of 15 such occurrences, 10 videos were marked as high-quality by all experts and only 5 videos were marked as low-quality by at least one expert.
\section{Discussion}

This paper makes three significant contributions. First, it demonstrates that the finger-tapping task can be reliably assessed by neurologists from remotely recorded videos. Second, it suggests that AI-driven models can perform close to clinicians and possibly better than non-specialists in assessing the finger-tapping task. Third, the model is equitable across gender, age, and PD vs. non-PD groups. These offer new opportunities for utilizing AI to address movement disorders, extending beyond Parkinson's disease to encompass other conditions like ataxia and Huntington's disease, where finger-tapping provides valuable insights into the severity of the disease.

Our tool can be expanded to enable longitudinal tracking of symptom progression to fine-tune the treatment of PD. People with PD (PwP) often exhibit episodic symptoms, and longitudinal studies require careful management of variables to ensure accurate temporal responses to individual doses of medication. It is best practice to conduct repeated ratings under consistent conditions, such as at the same time of day, the same duration after the last medication dose, and with the same rater \cite{perlmutter2009assessment}. However, the limited availability of neurological care providers and mobility constraints of elderly PwP make this challenging. In the future, we envision extending our platform for other neurological tasks (e.g., postural and rest tremors, speech, facial expression, gait, etc.) so that patients can perform an extensive suite of neurological tasks in their suitable schedule and from the comfort of their homes. For this use case, our tool is not intended to replace clinical visits for individuals who have access to them. Instead, the tool can be used frequently between clinical visits to track the progression of PD, augment the neurologists' capability to analyze the recorded videos with digital biomarkers, and fine-tune the medications. In healthcare settings with an extreme scarcity of neurologists, the tool can take a more active role by automatically assessing the symptoms frequently and referring the patient to a neurologist if necessary.

We introduce several digital biomarkers of PD -- objective features that are interpretable, clinically useful, and significantly associated with the clinical ratings. For example, the most significant feature correlated with the finger-tapping severity score is the inter-quartile range (IQR) of finger-tapping speed (Table \ref{tab:features}). It measures one's ability to demonstrate a range of speeds (measured continuously) while performing finger-tapping and is negatively correlated with PD severity. As the index finger is about to touch the thumb finger, one needs to decelerate and operate at a low speed. Conversely, when the index finger moves away from the thumb finger after tapping, one needs to accelerate and operate at high speed. A higher range implies a higher difference between the maximum and minimum speed, denoting someone having more control over the variation of speed and, thus, healthier motor functions. Moreover, the median and the maximum finger-tapping speed, as well as the median and the maximum finger-tapping amplitudes have strong negative correlations with PD severity. Finally, IQR and standard deviation of finger tapping frequency, as well as entropy, variance, and IQR of tapping periods, are found to have strong positive correlations with PD severity, as they all indicate the absence of regularity in the amplitude and periods. These findings align with prior clinical studies reporting that individuals with Parkinson's have a slower and less rhythmic finger tapping than those without the condition \cite{kim2011quantification}. These computational features can not only be used as \emph{digital biomarkers} to track the symptom progression of PwP but also explain the model's predicted severity score (e.g., an increase in the severity score can be attributed to factors like reduced tapping speed, smaller amplitude, etc.) However, some of the features examined in this study might be influenced by tremor, a significant symptom of Parkinson's disease that can frequently obscure signs of bradykinesia. For instance, accurately identifying individual finger taps necessitates precise peak detection from the finger-tapping motion over time. When tremors impact the motion, the signal can become unstable, posing a challenge in detecting clear and distinct peaks. Errors in peak detection would consequently impact the assessment of several features employed in this study, including finger-tapping period, frequency, and amplitude. Additionally, it is possible that severe tremors will affect the performance of pose estimation models. Pose estimation models operate by tracking the motion of body parts in a video. Tremor can induce erratic and unpredictable motion of the body parts, which can impede the model's ability to track the motion. This can result in inaccuracies in estimating the pose, ultimately compromising the accuracy of the extracted features. Unfortunately, we do not have the tremor diagnosis for the participants in this study and, therefore, could not provide definitive answers to these concerns. It is worth noting that, due to the jerky and unpredictable movements caused by tremor, doctors also encounter difficulties in assessing the speed of an individual's motion, rendering the diagnosis of bradykinesia challenging. Future studies may further investigate the connection between tremors and bradykinesia.

As we prepare to roll out our AI tool in healthcare settings, we must prioritize ethical considerations such as data security, user privacy, and algorithmic bias. As AI platforms become increasingly integrated into healthcare domains, there is growing emphasis on protecting against data breaches and crafting appropriate regulatory frameworks prioritizing patient agency and privacy \cite{murdoch2021privacy}. This ever-evolving landscape will have significant implications for our future approach. Additionally, algorithmic bias and its risks in perpetuating healthcare inequalities will present ongoing challenges \cite{mhasawade2021machine}. Many AI algorithms tend to underdiagnose the underserved groups \cite{seyyed2021underdiagnosis}, and it is critical to evaluate and report the model performance across age, race, and gender groups. Our proposed model does not demonstrate detectable bias across the male and female population, people with and without PD, white and non-white population, and to a particular age group. However, there is still room for improvement. Although not statistically significant, the model is slightly more inaccurate for the male (vs. female) and non-white (vs. white) populations. The higher average error for male can be explained by the fact that in our dataset, videos of male subjects had significantly higher severity\footnote{p-value obtained by one-tailed t-test = 0.003} (n = 267, mean severity = 1.43, std = 0.98) than female subjects (n = 222, mean severity = 1.19, std = 1.94) and, we had less data for modeling the higher severity classes. Also, higher error for non-white population could be due to having less data to represent them. Diversifying our training data and gathering feedback from critical stakeholders (especially from traditionally underrepresented and underserved communities) will be important first steps for us to take toward building fair, high-performance algorithms in the future. To that end, we will diversify our dataset to be representative of the general population. Notably, 92\% of participants in this study self-reported as white. Non-white races are typically underrepresented in clinical research \cite{clark2019increasing}. Thus, emphasizing recruitment of these populations through targeted outreach will be essential, especially considering the risks that homogeneous training data can pose in furthering healthcare inequalities \cite{gianfrancesco2018potential}.

The model exhibits an average prediction error of 0.58, indicating that it frequently predicts a level 1 severity for a healthy individual who actually has a severity score of 0. As observed in the confusion matrix, the model misclassifies 63\% of the ground-truth zero severity scores as severity 1. Moreover, it is less accurate for videos with severity 3 and 4. We achieved an overall accuracy of 50.92\% when utilizing the LightGBM regressor as a classifier, highlighting the need for further enhancement in the model. However, it is important to note that the lack of accuracy is also evident among experts and non-experts. On average, a pair of experts only concurred on the severity of a finger-tapping video 51.35\% of the time. Additionally, the non-experts obtained an overall accuracy of 36.03\%. This underscores the challenge in precisely assessing symptoms of Parkinson's disease. In clinical settings, although assessment of motor signs is important, it alone does not determine a Parkinson's disease diagnosis. Clinicians also consider the patient's health history, medications, and non-motor symptoms, such as anxiety, depression, impaired sense of smell, constipation, and changes in sleep among other factors, to determine a diagnosis. However, MDS-UPDRS scores are highly suitable for monitoring individuals already diagnosed with Parkinson's disease. An increase in the severity score from the baseline indicates the manifestation of more Parkinsonian symptoms, while a decrease suggests an improvement in symptoms. Considering this use case, severity assessment is typically regarded as a regression problem rather than a classification problem. Thus, having a strong correlation (e.g., Pearson's correlation coefficient) and low error (e.g., mean absolute error) with respect to the ground-truth labels are the most desirable metrics.

%%Discussion on video quality
When developing tools for analyzing data recorded in home environments, it is crucial to consider the various types of noise that can naturally occur in such settings. Home videos may exhibit background noise, inadequate lighting, blurriness, and other artifacts. These factors can pose challenges for both doctors and models in evaluating the videos. In this study, at least one of the experts expressed discomfort in rating 104 out of 489 videos due to quality issues. Although the difference was not statistically significant, there was a slightly lower level of inter-rater agreement among the experts when it came to the low-quality videos. However, the model's performance, as indicated by the prediction error, was similar for both good and poor quality videos. It is possible that the video quality remained sufficient for the pose-estimation model and the feature-extraction framework employed to automatically assess the severity of the finger-tapping task. Nevertheless, further analysis indicates that MediaPipe (the pose-estimation model used in this study) struggles when significant external systematic noise is introduced into the videos. For instance, the confidence of MediaPipe hand-tracking dropped when a subset of the good-quality videos was intentionally blurred or when random Gaussian noise was added. In general, we recommend ensuring a minimum level of video quality to obtain reliable ground truth and facilitate the use of more precise tools for video processing. The proposed AI-driven model (i.e., LightGBM regressor) was trained with 489 videos from 250 global participants. While this dataset is the largest in the literature in terms of unique individuals, it is still a relatively small sample size for training models capable of capturing the essence of complex diseases such as Parkinson's. Furthermore, it is worth noting that the dataset utilized in this study exhibits class imbalance. More specifically, there is a scarcity of samples for severity classes 3 and 4. While the number of videos for severity classes 0, 1, and 2 amounted to 108, 181, and 141 respectively, there were only 54 and 5 videos available for severity classes 3 and 4. This class imbalance may have contributed to the model's inability to predict the most severe class (i.e., severity 4) correctly. In the future, we plan to improve our model's performance by building (i) a larger dataset with a better balance in severity scores and (ii) a gatekeeper to improve data quality. Additional data will be essential in building more powerful models with potentially better performance. Furthermore, we can improve our data quality and model performance by developing ``quality control'' algorithms to provide users with real-time feedback on capturing high-quality videos: such as adjusting their positioning relative to the webcam or moving to areas with better lighting. Developing user-friendly features to strengthen data collection and ensure minimum video quality will be crucial for collecting videos remotely without direct supervision. Additionally, it is worth noting that the symptoms of Parkinson's disease can vary depending on whether an individual is in the ON-state (under the effect of PD medication) or in the OFF-state (not under the effect of PD medication). It would be intriguing to investigate whether the model can effectively detect differences in symptom severity between participants who are ON or OFF PD medication in future studies.
\section{Methods}
\subsection*{Data Sources}
Participants' data were collected through a publicly accessible web-based tool\footnote{\url{https://parktest.net/}}. This tool allows individuals to contribute data from the comfort of their homes, provided they have a computer browser, internet connection, webcam, and microphone. In addition to the finger-tapping task, the tool also gathers self-reported demographic information such as age, gender, race, and whether the participant has been diagnosed with Parkinson's disease (PD) or not. Moreover, the tool records other standard neurological tasks involving speech, facial expressions, and motor functions, which can help to extend this study in the future.

We collected data from 250 global participants who recorded themselves completing the finger-tapping task in front of a computer webcam. Data was collected primarily at participants' homes; however, a group of individuals (48) completed the task in a clinic using the same web-based tool. Study coordinators were available for the latter group if the participants needed help. The demographic characteristics of the study participants is provided in Table \ref{tab:demography}.

\subsection*{Clinical Ratings}
The finger-tapping task videos were evaluated by a team of five raters, including two non-specialists and three expert neurologists. The expert neurologists are all associate or full professors in the Department of Neurology at a reputable institution in the United States, possess vast experience in PD-related clinical studies, and actively consult PD patients. Both of the non-specialists are MDS-UPDRS certified independent raters. One of the non-specialists holds a non-U.S. bachelor's degree in medicine (MBBS) and has experience conducting PD clinical studies. The other non-specialist is a second-year neurology resident with 10 years of experience in clinical research related to movement disorders. The first non-specialist does not consult PD patients and the second non-specialist takes care of less than 20 PD patients per year.

The raters watched the recorded videos of each participant performing the finger-tapping task and rated the severity score for each hand following the MDS-UPDRS guideline\footnote{\url{https://www.movementdisorders.org/MDS-Files1/Resources/PDFs/MDS-UPDRS.pdf}} (Part III, Section 3.4). The severity rating is an integer ranging from 0 to 4 representing normal (0), slight (1), mild (2), moderate (3), and severe (4) symptom severity. The rating instructions emphasize focusing on speed, amplitude, hesitations, and decrementing amplitude while rating the task. In addition to providing the ratings, the raters could also mark videos where the task was not properly performed or when a video was difficult to rate. We excluded the difficult-to-rate videos marked by any of the experts when analyzing the performance of the raters.

To compute the ground-truth severity scores, we considered only the ratings the three expert neurologists provided. If at least two experts agreed on the severity rating for each recorded video, this was recorded as the ground truth. If the experts had no consensus, their average rating rounded to the nearest integer was considered the ground truth. The ratings from the non-specialists were used solely to compare the machine-learning model's performance.

\subsection*{Feature Extraction}
We developed a set of features by analyzing the movements of several key points of the hand. The feature extraction process is comprised of four stages: (i) distinguishing left and right-hand finger-tapping from the recorded video, (ii) quantifying finger-tapping movements by extracting key points on the hand, (iii) reducing noise, and (iv) computing features that align with established clinical guidelines, such as MDS-UPDRS.

\subsubsection{Hand separation:} The finger-tapping task is performed for both hands, one hand at a time. However, to rate each hand independently, we divided the task video into two separate videos, one featuring the right hand and the other featuring the left hand. We manually reviewed each video and marked the transition from one hand to the other. The data collection framework will be designed to record each hand separately to avoid manual intervention in the future.

\subsubsection{Extracting hand key points:} After separating the left and right-hand finger-tapping videos, we applied MediaPipe Hands \footnote{\url{https://google.github.io/mediapipe/solutions/hands.html}} to detect the coordinates of 21 key points on each hand. MediaPipe is an open-source project developed by GoogleAI that provides a public API of a highly accurate state-of-the-art model for hand pose estimation. In addition, the hand pose estimation model is very fast, easy to integrate into a machine learning framework, and supports various platforms, including Android, iOS, and desktop computers. Furthermore, GoogleAI consistently updates the pose estimation models and seamlessly integrates these updates into the public API. Consequently, we selected MediaPipe over other pose estimation platforms due to these compelling reasons. Among the MediaPipe extracted hand key points, we utilized the \texttt{thumb-tip}, \texttt{index-finger-tip}, and the \texttt{wrist}, which are critical to track the finger-tapping task-specific movements. In cases where multiple persons are in the frame (i.e., one in the background), we discarded the smaller hand's key points as the individual performing the task was typically closer to the camera, resulting in a larger hand appearance in the recorded video. For the left-handed finger-tapping video, we only tracked the key points on the left hand, and similarly, for the right-handed videos, we only tracked the key points on the right hand. Instead of using Euclidean distance between the thumb-tip and index finger-tip to measure the amplitude, speed, and other metrics to quantify finger-tapping movements, we used the angle incident by three key points: thumb-tip, wrist, and index finger-tip. This helped us to deal with participants sitting at a variable distance from the camera since the angle is invariant to the camera distance. We computed the angle for each frame of the recorded video (i.e., if a video was recorded at 30 frames/second, we computed the angle 30 times per second). This helped us assess the speed and acceleration of the fingers in a continuous manner.

\subsubsection{Noise reduction:} 

\begin{figure*}
\centering

\includegraphics[width=\linewidth]{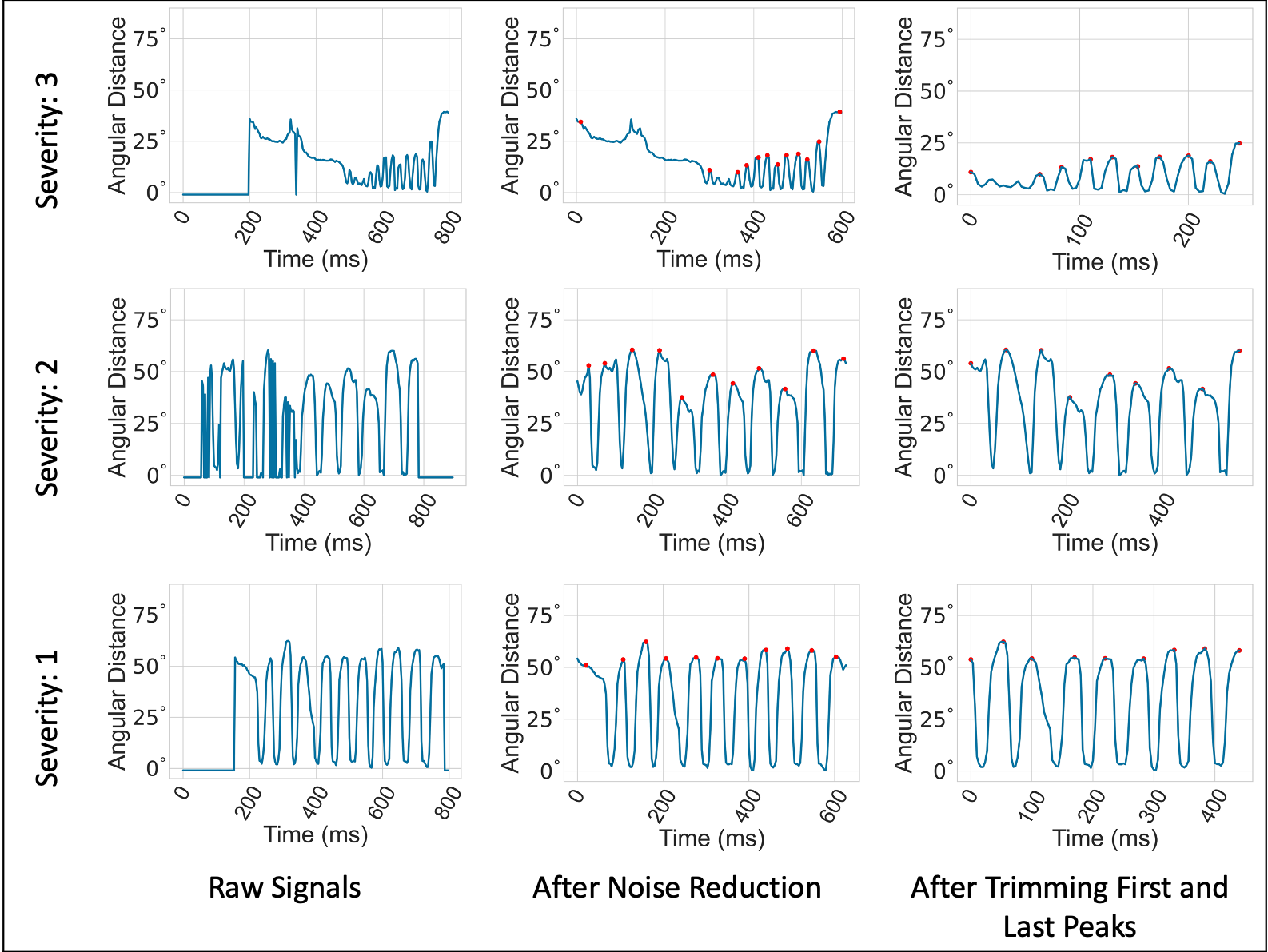}
%\caption{Instead of using complex models that try to fuse multiple modalities, we rely on a single pre-trained language model. As input to language model, we extend the utterance text with visual-text and acoustic-text extracted from communication behavior videos. We group the nonverbal (visual and acoustic) features that frequently appear together into clusters, and use textual descriptions of the clusters to extract these nonverbal-texts.\commentG{rearrange the sentences}}
\caption{\textbf{Data pre-processing.} Finger-tapping angles incident by three hand key points (thumb-tip, wrist, index finger-tip) plotted as a time series. Figures on the left show the noisy raw signals directly extracted using MediaPipe. After the noise reduction step, we identified peak angles (red dots) using a custom peak detection algorithm. Finally, trimming the signal by removing the first and last tap yields the cleanest signal used for analysis, as shown on the right. The top figures depict a person with severe tapping difficulty (severity: 3), resulting in low and irregular amplitudes. The central figures show a person with moderate tapping ability (severity: 2), with slow and interrupted tapping and irregular amplitudes. Finally, the bottom figures show a person with good rhythmic tapping ability, albeit with a slower tapping speed (severity: 1).}%
\label{fig:noise}
\end{figure*}

The computed finger-tapping angles are plotted as a time-series signal in Figure \ref{fig:noise} (left). Negative values of the angle are indicative of a missing hand in the captured frame. For example, at the beginning or at the end of the recording, the hand might not be visible in the recording frame, as the participant needs to properly position their hand. We detected missing hands using the handedness outputs from MediaPipe, which include information about the left/right hand and hand presence confidence score. In particular, if the target hand (left/right) is not recognized as a handedness category or if the hand presence confidence score is less than 0.90, we labeled the frame as having a missing hand. In these cases, we assign a finger-tapping angle of -1.0. However, MediaPipe can also inaccurately miss the hand in many frames, resulting in a negative value for the angle. To address this issue, we implemented a strategy to interpolate missing angle values when the majority of neighboring frames have non-negative values. Specifically, we looked at the five frames before and after the missing value. If the majority had non-negative angles, we interpolated the value using a polynomial fit on the entire signal. Then, we found the largest consecutive segment of frames where the hand is visible (i.e., the finger-tapping angle was non-negative) in the signal and remove the frames before and after that. This helps us to remove the pre and post-task segments where the participants were not tapping their fingers and ensure that the analysis is focused on the relevant segment of the signal. Figure \ref{fig:noise} (middle) shows how the raw, noisy signals were converted to cleaner signals after performing this step.

The participants need to adjust the positioning of their hands before starting to tap their fingers, and they also need to move their hands after completing the task, which introduces further noise to the signal. Specifically, it can impact the first and last tap they undergo. As the task instructs the participants to tap their fingers ten times, we decided to remove the first and last tap, hoping to obtain the cleanest signal to analyze. To accomplish this, we ran a custom peak detection algorithm to find the peaks of the finger-tapping angle, and we removed the portion of the signal before the second peak and after the second-last peak. The peak detection algorithm utilizes some of the unique properties of the task. For example, a peak must be followed by a bottom (i.e., low value of finger-tapping angle) as the tapping is repetitive, the duration between two subsequent taps cannot be too small (i.e., determined by the fastest finger-tapping speed recorded), and the peaks must be bigger than the smallest 25 percentile values of the signal. Figure \ref{fig:noise} (right) demonstrates the effectiveness of this step, as it helps to obtain a clean signal that can be used to develop objective measures of the finger-tapping task.

\subsubsection{Computational features:} The coordinates of the wrist, thumb-tip, and index finger-tip, and their incident angle were used to compute the following features:

\begin{itemize}
    \item \textbf{Finger-tapping period} is measured as the time (in seconds) it took for a participant to complete a tap.

    \item \textbf{Finger-tapping frequency} is the inverse of the finger-tapping period, measuring the number of taps completed per second.

    \item \textbf{Amplitude} is defined as the maximum angle (in degree) made by the thumb-tip, wrist, and index finger-tip while completing a tap. As explained earlier, an angular measurement is more reliable than linear (i.e., Cartesian) distance since angles are invariant to distance from the camera.

    \item \textbf{Finger-tapping speed} is a continuous measurement of an individual's tapping speed. For each recorded frame, we quantify the change in the finger-tapping angle compared to the previous frame and multiply this with the recorded video's frame rate (so the speed is measured in degree/second unit). The average frame rate of the recorded videos was 30, meaning that we can measure an individual's tapping speed 30 times a second.

    \item \textbf{Acceleration} is also a continuous measurement, which is the derivative of finger-tapping speed. Specifically, for each frame, we measure the change in speed compared to the previous frame to quantify acceleration in degree/second-square unit.

    \item \textbf{Wrist movement} measures the stability of the hand when performing the task. Ideally, the hand of a healthy individual should remain stable, and the wrist should not move much. However, people with PD may experience hand tremors and therefore demonstrate wrist movement. We compared the wrist's current coordinates for each frame with the previous frame's coordinates. We measured the absolute values of the movements along both the X and Y axes separately and also measured the Cartesian distance. All values were normalized by the distance between the wrist and the thumb carpometacarpal (CMC) to account for the distance from the camera.
\end{itemize}

For each of the features above, we measured the statistical median, inter-quartile range (IQR), mean, minimum, maximum, standard deviation, and entropy \footnote{Entropy is a measure of uncertainty or randomness in a signal, calculated using Shannon's formula.}, and used them as separate features. Period, frequency, and amplitude were measured discretely (i.e., for each tap), while speed, acceleration, and wrist movement were assessed continuously (i.e., at each frame). The detected peaks were used to separate each tap. 

Additionally, we measured the following aggregate features based on the entire signal to capture the rhythmic aspects of the finger-tapping task:

\begin{itemize}
    \item \textbf{Aperiodicity:} Periodicity is a concept borrowed from signal processing that refers to the presence of a repeating pattern or cycle in a signal. For example, a simple sinusoidal signal (e.g., $f(t) = sin(t)$) will have higher periodicity compared to a signal that is a combination of several sinusoidal signals (e.g., $f(t) = sin(t) + sin(2t)$). Aperiodicity measures the absence of periodicity (i.e., the absence of repeating patterns). To measure aperiodicity, signals are transformed into the frequency domain using Fast Fourier Transformation (FFT). The resulting frequency distribution can be used to calculate the normalized power density distribution, which describes the energy present at each frequency. The entropy of the power density distribution is then computed to measure the degree of aperiodicity in the signal. A higher entropy value indicates greater uncertainty in the frequency distribution, which corresponds to a more aperiodic signal. A similar measure was found to be effective in measuring the symptoms of Alzheimer's disease \cite{sharma2020biophysical}.
    
    \item \textbf{Number of interruptions:} Interruption is defined as the minimal movement of the fingers for an extended duration. We calculated a distribution of continuous finger-tapping speeds across the study population. Our analysis revealed that over 95 percent of the tapping speeds exceeded 50 degrees/second. As a result, any instance where an individual's finger-tapping speed was less than 50 degrees/second for at least ten milliseconds (ms) was marked as an interruption. The total number of interruptions present in the recorded video was then computed using this method.
    
    \item \textbf{Number of freezing:} In our study, we considered freezing as a prolonged break in movement. Specifically, any instance where an individual recorded less than 50 degrees/second for over 20 ms was identified as a freezing event, and we counted the total number of such events.
    
    \item \textbf{Longest freezing duration:} We recorded the duration of each freezing event and calculated the longest duration among them.
    
    \item \textbf{Tapping period linearity:} We recorded the tapping period for each tap and evaluated the possibility of fitting all tapping periods using a linear regression model based on their degree of fitness ($R^2$). Additionally, we determined the slope of the fitted line. The underlying idea was that if the tapping periods were uniform or comparable, a straight line with slope=0 would adequately fit most periods. Conversely, a straight line would not be an appropriate fit if the periods varied significantly.

    \item \textbf{Complexity of fitting periods:} The complexity of fitting finger-tapping periods can provide insights into the variability of these periods. To measure this complexity, we used regression analysis and increased the degree of the polynomial from linear (degree 1) up to 10. We recorded the minimum polynomial degree required to reasonably fit the tapping periods (i.e., $R^2 \ge 0.9$).
    
    \item \textbf{Decrement of amplitude:} Decrement of amplitude is one of the key symptoms of Parkinsonism. We measured the finger-tapping amplitude for each tap and quantified how the amplitude at the end differed from the mean amplitude and amplitude at the beginning. Additionally, we calculate the slope of the linear regression fit to capture the overall change in amplitude from start to end.
\end{itemize}

\subsubsection{Highly-correlated feature removal and significance test:} The abovementioned measurements and some of their statistical aggregates result in 65 features used to analyze the recorded finger-tapping videos. We perform a cross-correlation analysis among the features, identify the highly correlated pair (i.e., Pearson's correlation coefficient, $r > 0.85$), and drop one from each pair. This helps to remove redundant features and enables learning the relationship between the features and the ground truth severity scores using simple models. This is important, as simple models tend not to over-fit the training data and are more generalizable than complex models (commonly known as Occam's razor \cite{blumer1987occam}). After this step, the number of features was reduced to 53.

For each of the 53 features, we perform a statistical correlation test to identify the features significantly correlated with the ground-truth severity score. Specifically, for each feature, we take the feature values for all the recorded videos in our dataset and the associated ground-truth severity scores obtained by the majority agreement of three expert neurologists. We measure the Pearson's correlation coefficient ($r$) between the feature values and severity scores and test the significance level of that correlation (i.e., $p$-value). We found 18 features to be significantly correlated (at a significance level, $\alpha = 0.01$). The significant features include finger-tapping speed (inter-quartile range, median, maximum, minimum), acceleration (minimum), amplitude (median, maximum), frequency (inter-quartile range, standard deviation), period (entropy, inter-quartile range, minimum), number of interruptions, number of freezing, longest freezing duration, aperiodicity, the complexity of fitting periods, and wrist movement (minimum Cartesian distance). Table \ref{tab:features} reports the correlation's direction, strength, and statistical significance level for the most correlated ten features. It is important to acknowledge that certain features may exhibit a strong non-linear correlation that is not adequately captured by Pearson's correlation coefficient. Due to this reason, we retained all 53 features as candidates for training machine learning models, even if some of them did not exhibit significant correlations.

\subsection*{Feature Processing, Model Training, Evaluation, and Explanation}

\subsubsection*{Feature selection:}
With a small dataset, there is an increased risk of overfitting, where the model learns the noise and specific patterns of the training data rather than generalizing well to unseen data. Feature selection helps mitigate this risk by reducing the complexity of the model and focusing on the most informative features, which reduces the likelihood of overfitting. We used the BoostRFE (Boosted Recursive Feature Elimination) method implemented in the shap-hypertune Python package with the LightGBM base model to reduce the number of features fed to the machine learning model. BoostRFE combines the concepts of boosting and recursive feature elimination. It identifies and ranks the most informative features in a dataset. After selecting the feature set, we scaled all the features based on training data to bring them onto the same scale. We used the ``number of top features'' to be selected by the BoostRFE method and the scaling method as hyper-parameters, and the best model picked the top-22 features out of 53 candidate features and StandardScaler (implemented in Python sklearn package) as the scaling method. Further details are available in the supplementary information.
\color{black}

\subsubsection{Model training:}
To model our dataset, we applied a standard set of regressor models (i.e., XGBoost, LightGBM, Support Vector Regression (SVM), AdaBoost, and RandomForest), as well as shallow neural networks (with 1 and 2 trainable layers). We run an extensive hyper-parameter search for all of these models using the Weights and Biases tool\footnote{\url{https://wandb.ai}} and found LightGBM \cite{ke2017lightgbm} to be the best-performing model (Table \ref{tab:performance_models}). LightGBM is a gradient boosting framework like XGBoost \cite{chen2016xgboost} that works by iteratively building a predictive model using an ensemble of decision trees. It uses a leaf-wise tree growth strategy where each tree is grown by splitting the leaf that offers the most significant reduction in the loss function. LightGBM is a popular choice in modeling structured data due to its fast training speed, low memory usage, and high performance. To run the hyper-parameter search for the LightGBM regressor, we experimented with different learning rates, maximum depth of the tree, number of estimators to use, etc. The list of all hyper-parameters, their search space, and the corresponding value for the best model is reported in the supplementary information.

\begin{table*}[]
\begin{tabular}{|l|l|l|l|l|l|l|l|}
\hline
\textbf{Model}                                                                               & \textbf{MAE}    & \textbf{MSE}    & \textbf{Accuracy} & \textbf{Kendall's Tau} & \textbf{MAPE}    & \textbf{PCC}    & \textbf{Spearman's $\rho$} \\ \hline
SVR                                                                                          & 0.5861          & 0.5586          & \textbf{51.94\%}  & 0.5044                & 32.14\%          & 0.6388          & 0.6329                \\ \hline
Random Forest Regressor                                                                      & 0.5920          & 0.5482          & 49.28\%           & 0.5116                & 33.5\%           & 0.6518          & 0.6389                \\ \hline
AdaBoost Regressor                                                                           & 0.5926          & 0.5499          & 45.6\%            & 0.5038                & 33.75\%          & 0.6219          & 0.6317                \\ \hline
XGBoost Regressor                                                                            & 0.5904          & 0.5553          & 51.33\%           & 0.4989                & 32.57\%          & 0.6417          & 0.6282                \\ \hline
\textbf{LightGBM Regressor}                                                                           & \textbf{0.5802} & \textbf{0.5364} & 50.92\%           & \textbf{0.5147}       & \textbf{32.01\%} & \textbf{0.6563} & \textbf{0.6429}       \\ \hline
\begin{tabular}[c]{@{}l@{}}Shallow Neural Network - I\\ (one trainable layer)\end{tabular}   & 0.6154          & 0.6007          & 46.83\%           & 0.4810                & 33.9\%           & 0.6097          & 0.6100                \\ \hline
\begin{tabular}[c]{@{}l@{}}Shallow Neural Network - II\\ (two trainable layers)\end{tabular} & 0.6069          & 0.6162          & 51.33\%           & 0.4813                & 32.42\%          & 0.6004          & 0.6044                \\ \hline
\end{tabular}
\caption{\textbf{Performance of different regressor models we tested.}}
\label{tab:performance_models}
\end{table*}

\subsubsection{Data imbalance:}
The dataset we analyzed exhibited class imbalance. Specifically, we had 108, 181, and 141 videos with severity scores of 0, 1, and 2 respectively, while the number of videos with severity scores 3 and 4 was significantly lower, with only 54 and 5 videos respectively. To address this issue, we experimented with a technique called SMOTE (Synthetic Minority Over-sampling Technique proposed by \cite{chawla2002smote}) proposed by Chawla et al. (2002). SMOTE generates synthetic data samples for the minority classes by selecting an instance from each minority class and choosing one of its $k$-nearest neighbors (where $k$ is a user-defined parameter). It then creates a synthetic instance by interpolating between the selected instance and the chosen neighbor. However, the model performance degraded (i.e., Pearson's correlation coefficient decreased from 0.6563 to 0.6422, and mean absolute error increased from 0.5802 to 0.5807) after integrating SMOTE. Therefore, we ended up not using minority oversampling.

% We developed a machine learning model (i.e., XGBoost regressor \cite{chen2016xgboost}) that predicts the severity of PD symptoms based on the extracted features from a finger-tapping video.
\subsubsection{Evaluation:}
To assess the model's performance, we employed a leave-one-patient-out cross-validation (LOPO-CV) technique. LOPO-CV involves partitioning the dataset in a manner where each patient's data is treated as a distinct validation set, while the remaining data is utilized for training the model. This process entails multiple iterations, with each iteration excluding the data samples of a specific patient (both left and right hand videos in our case) as the test set, while the machine learning model is trained on the remaining data. LOPO-CV ensures that the model's performance is evaluated on unseen patients, resembling real-world scenarios where the model encounters new patients during deployment. This approach is particularly well-suited for machine learning applications in the healthcare domain.

%To evaluate the model, we implemented a stratified $k$-fold cross-validation technique (with $k=10$). Stratified cross-validation is similar to k-fold cross-validation, but it ensures that each fold has approximately the same proportion of target values as the entire dataset. This is particularly useful when the target variable is imbalanced, meaning that there are significantly more instances of one class than another. Stratified cross-validation can help ensure that each fold has a representative sample of both target classes, which can help prevent the overestimation of model performance on the majority class. For each combination of hyper-parameter choices, we ran stratified $k$-fold cross-validation for 20 random iterations. 
We used seven metrics to evaluate the performance of several regression models attempted to measure the severity of the finger-tapping task: mean absolute error (MAE), mean squared error (MSE), classification accuracy, Kendall rank correlation coefficient (Kendall's Tau), mean average percentage error (MAPE), Pearson's correlation coefficient (PCC), and Spearman's rank correlation coefficient (Spearman's $\rho$). MAE is a metric used to measure the average magnitude of the errors in a set of predictions without considering their direction. It is calculated by taking the absolute differences between the predicted and actual values and then averaging those differences. The smaller the MAE, the better the model is performing. MSE is also a measure of model's error, however, it penalizes the bigger errors more as the errors are squarred. Instead of capturing errors in an absolute scale, MAPE measures errors in a relative scale. For example, making an one-point error when the ground-truth value is 4 will be considered as 25\% error using MAPE. Both Kendall's Tau and Spearman's $\rho$ is popularly used in statistics to measure the ordinal association between two measured quantities. PCC measures the strength and direction of the relationship between two variables and ranges from -1 to +1. A correlation of -1 indicates a perfect negative relationship, a correlation of +1 indicates a perfect positive relationship, and 0 indicates no relationship between the variables. Finally, accuracy captures the percentage of time the model is absolutely correct (when the regression values are converted to five severity classes). Note that, unlike the other metrics, accuracy does not distinguish between making a one-point error and larger errors, and thus it is not commonly used in regression tasks. In general, these broader sets of metrics provide a more detailed and diverse picture of the model performance.

\subsubsection{Model interpretation:}
To explain the model's performance, we used SHAP. SHAP (SHapley Additive exPlanations) is a tool used for explaining the output of any supervised machine learning model~\cite{lundberg2020local2global}. It is based on Shapley values from cooperative game theory, which allows us to assign an explanatory value to each feature in the input data. The main idea behind SHAP is to assign a contribution score to each feature that represents its impact on the model's output. These contribution scores are calculated by considering each feature's value in relation to all possible combinations of features in the input data. By doing this, SHAP can provide a detailed and intuitive explanation of why a particular model made a certain decision. This makes it easier for human decision-makers to trust and interpret the output of a machine-learning model.

% Additionally, the continuous prediction is converted to four severity classes (0, 1, 2, and 3) by rounding the prediction to the closest integer. The classification result is reported in Figure \ref{fig:ml_performance},

\subsection*{Use of Large Language Models}
ChatGPT\footnote{\url{https://chat.openai.com/chat}} -- a large language model developed by OpenAI\footnote{\url{https://openai.com/}} that can understand natural language prompts and generate text -- was used to edit some part of the manuscript (i.e., suggest improvements to the language, grammar, and style). All suggested edits by ChatGPT were further verified and finally integrated into the manuscript by an author. Please note that ChatGPT was used only to suggest edits to existing text, and we did not use it to generate any new content for the manuscript.

\subsection*{Ethics}
The study was approved by the institutional review board (IRB) of the University of Rochester, and the experiments were carried out following the approved study protocol. We do not have written consent from the participant as the study was primarily administered remotely. However, participants provided informed consent electronically for the data used for analysis and photos presented in the figures.

\section*{Code and Data Availability}
The recorded videos were collected using a web-based tool. The tool is publicly accessible at \textcolor{blue}{\url{https://parktest.net}}. The codes for video processing and feature extraction, as well as the trained model, will be made publicly available upon the acceptance of this paper. We will provide a link to the repository containing the codes and models in the paper.

Unfortunately, we are unable to share the raw videos due to the Health Insurance Portability and Accountability Act (HIPAA) compliance. However, we are committed to sharing the extracted features upon receiving an email request at \textcolor{blue}{\url{rochesterhci@gmail.com}}. The features will be provided in a structured format that can be easily integrated with existing machine-learning workflows.

For potential collaboration, we welcome interested individuals or groups to reach out to us at \textcolor{blue}{\url{mehoque@cs.rochester.edu}}. Depending on the specifics of the collaboration, we may be able to share some additional data beyond the extracted features.

\bibliography{main}

\section*{Acknowledgement}
This work was supported by the U.S. Defense Advanced Research Projects Agency (DARPA) under grant W911NF19-1-0029, National Science Foundation Award IIS-1750380, National Institute of Neurological Disorders and Stroke of the National Institutes of Health under award number P50NS108676 and Moore Foundation.

\section*{Competing Interests}
The authors declare that there are no competing interests.

\section*{Author Contributions}
M.S.I., W.R., P.T.Y., J.L.P., J.L.A., R.B.S., E.R.D., and E.H. conceptualized and designed the experiments. A.A. and S.L. implemented the web-based data collection framework, M.S.I. designed and implemented the feature extraction techniques, and W.R. worked on the machine learning model. M.S.I. worked on data analysis and visualization. M.S.I., W.R., P.T.Y., and E.H. wrote the manuscript. All the authors read the manuscript and provided valuable suggestions for revising it. All authors accept the responsibility to submit it for publication.

\end{document}

\pagestyle{plain}
\maketitle

\section*{Supplementary Note 1 -- Feature Extraction}

The input to the feature extraction pipeline is a video $V$ (of the finger-tapping task) and a hand category $h$ (i.e., left/right). If the hand category is specified as left, the analysis focuses solely on the movements of the left hand, and the same applies to the right hand category. 

\subsection*{Locating target hand}
Initially, we processed the finger-tapping task video $V$ in a frame-by-frame manner using MediaPipe. For each frame $V_i$, we first tried to locate the specified hand category $h$. Using the \texttt{multi\_handedness} output from MediaPipe, we can identify how many different hands MediaPipe has detected and the handedness category and confidence score of detection for each detected hand. To identify the detected hand(s) that match the specified hand category $h$, we used the following heuristics:

\begin{algorithm}
\begin{algorithmic}
    \State $\text{hands\_found} \gets [\text{ }]$
    \For{all $j \text{ in range } (len(\text{multi\_handedness}))$}
        \If{(multi\_handedness[$j$].classification[0].label $ = h$) \& (multi\_handedness[$j$].classification[0].score $> 0.9$)}
            \State $\text{hands\_found}.append(j)$
        \EndIf    
    \EndFor
\end{algorithmic}
\end{algorithm}

\noindent We identified three possible scenarios:

\begin{itemize}
    \item \textbf{MediaPipe did not detect any hand with matching hand category.} In this case, we simply assigned the frame to have missing hand.
    \item \textbf{MediaPipe detected a single hand matching the hand category.} We considered this to be the hand we will analyze.
    \item \textbf{Mediapipe detected multiple hands matching the hand category.} When multiple persons are visible in the frame, MediaPipe can detect multiple right hands or multiple left hands. In such situations, we made an assumption that the person performing the task will have a larger hand compared to the other individual(s) in the background, as they are likely to be closest to the camera. To identify the largest hand, we compared the Euclidean distance between the coordinates of the wrist (\texttt{landmark[0]}) and the thumb-tip (\texttt{landmark[4]}) of the detected hands. The hand with the greatest distance was selected for further analysis.
\end{itemize}

\subsection*{Computation of features from MediaPipe outputs}
After we identified the target hand from the MediaPipe detected hands for frame $V_i$, we extracted the \texttt{x} and \texttt{y} coordinates of four landmarks (i.e., hand key points): centre of wrist joint (\texttt{WRIST, landmark[0]}), thumb carpometacarpal joint (\texttt{THUMB\_CMC, landmark[1]}), tip of the thumb (\texttt{THUMB\_TIP, landmark[4]}), and tip of the index finger (\texttt{INDEX\_FINGER\_TIP, landmark[8]}). These coordinates were used to track the finger-tapping angle and movement of wrist over time.

\textbf{wrist movement features:} To track wrist movement throughout the duration of the task, we recorded $W_i = (W_i.x, W_i.y)$, the x and y coordinates of the \texttt{WRIST} for each frame $V_i$. Note that, to account for the variable distance of the hand from the camera, the coordinates were normalized by the Euclidean distance between the \texttt{WRIST} and \texttt{THUMB\_CMC}.

For the $i^{th}$ frame ($i>1)$, we computed three metrics for capturing the wrist movement: 
\begin{itemize}
    \item absolute movement along $X$-axis, $$\Delta WX_i = |W_i.x - W_{i-1}.x|$$
    \item absolute movement along $Y$-axis, $$\Delta WY_i = |W_i.y - W_{i-1}.y|$$
    \item Cartesian distance, $$\Delta W_i = \sqrt{(W_i.x - W_{i-1}.x)^2 + (W_i.y - W_{i-1}.y)^2}$$
\end{itemize}

We finally computed statistical aggregates (i.e., median, inter-quartile range (IQR), mean, minimum, maximum, standard deviation, and entropy) to extract the wrist movement features.

\begin{figure}[H]
\centering

\includegraphics[width=\linewidth]{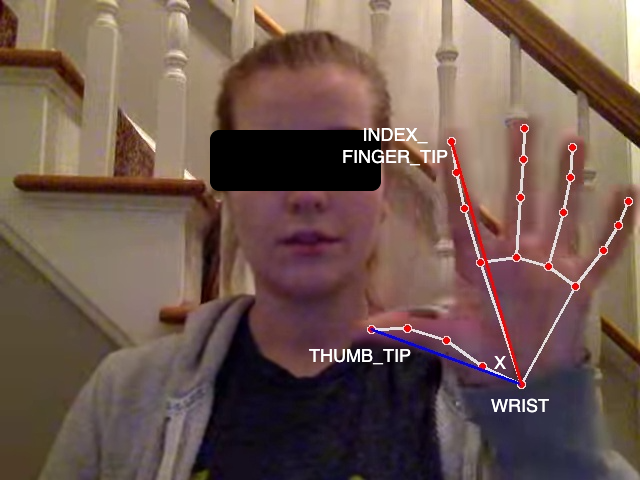}
\caption{\textbf{Hand key points extracted by MediaPipe.} All the extracted key-points are displayed as red dots. The three key-points WRIST, THUMB\_TIP, and INDEX\_FINGER\_TIP were used to compute the finger-tapping angle X.}
\label{fig:angle}
\end{figure}

\textbf{features based on finger-tapping angle:} We used the \texttt{WRIST}, \texttt{THUMB\_TIP}, and \texttt{INDEX\_FINGER\_TIP} coordinates to compute the finger-tapping angle $X_i$ for the $i^{th}$ frame as shown in Figure \ref{fig:angle}. Specifically, for this frame, we first identified two vectors $\overrightarrow{WT_i}$ and $\overrightarrow{WI_i}$, which represent the lines connecting the wrist to the thumb-tip and the wrist to the index finger-tip, respectively. The finger-tapping angle $X_i$ was then computed using the following formula:

$$X_i = \cos^{-1} \left(\frac{\overrightarrow{WT_i} \cdot \overrightarrow{WI_i}}{|\overrightarrow{WT_i}| \times |\overrightarrow{WI_i}|}\right)$$

In this context, the symbol $\cdot$ represents the dot product operation performed on two vectors, while $|\text{ }|$ denotes the magnitude of a vector.

The computed finger-tapping angles were then regarded as a time-series ($X_1, X_2, \cdots, X_n$ where $n$ is the number of frames in the video), which was further processed to reduce noise (please see the Noise reduction step in the ``Methods'' section.) Let $t_{frame}$ be the average duration of one frame for the video being analyzed (computed by dividing the entire video duration by the number of frames in the video). From the entire time-series, we computed the following metrics for each frame $V_i$ (where $i > 1$):

\begin{itemize}
    \item finger-tapping speed, $$v_i = \frac{|X_i - X_{i-1}|}{t_{frame}}$$

    \item acceleration, $$a_i = \frac{|v_i - v_{i-1}|}{t_{frame}}$$
\end{itemize}

In order to obtain additional important metrics, we utilized a custom peak detection algorithm to identify the peaks in the finger-tapping angles throughout the duration of the video. Let's denote the frame numbers at which the peaks occur as $P_1, P_2, \cdots, P_k$, where $k$ represents the total number of peaks in that specific video. Consequently, the corresponding peak values of the finger-tapping angles can be represented as $X_{P_1}, X_{P_2}, \cdots, X_{P_k}$. Due to the periodic and repetitive nature of the finger-tapping task, each peak can be used to separate one tap from the others. Utilizing these peak values, we proceeded to calculate the following metrics at each peak $P_i$ (where $i > 1$):

\begin{itemize}
    \item finger-tapping period, $$T_{P_i} = (P_{i} - P_{i-1}) \times t_{frame}$$
    \item finger-tapping frequency, $$f_{P_i} = \frac{1}{T_{P_i}}$$
    \item amplitude, $$A_{P_i} = X_{P_i}$$
\end{itemize}

Please note that while speed and acceleration were computed for each frame, the other metrics were computed for each detected peak. Statistical aggregates (median, inter-quartile range (IQR), mean, minimum, maximum, standard deviation, and entropy) of these metrics were used as features that are input to the machine learning model. In addition to these statistical aggregate features, we computed features like aperiodicity, number of interruptions, number of freezing, longest freezing duration, tapping period linearity, complexity of fitting periods, and decrement of amplitude by analyzing the entire time series of finger-tapping angles as a whole (please see the ``methods'' section of the main article for details).

\section*{Supplementary Note 2 -- Package Versions}

We used Python to conduct the experiments. The versions for each package used are listed below:
\begin{itemize}
    \item click=8.1.3
    \item cuda-cudart=11.8.89
    \item cuda-cupti=11.8.87
    \item cuda-libraries=11.8.0
    \item cuda-nvrtc=11.8.89
    \item cuda-nvtx=11.8.86
    \item cuda-runtime=11.8.0
    \item imbalanced-learn=0.10.1
    \item lightgbm=3.3.5
    \item matplotlib=3.7.1
    \item mediapipe=0.8.10
    \item numba=0.57.0
    \item numpy=1.24.3
    \item pandas=2.0.2
    \item pip=22.3.1
    \item python=3.9.13
    \item pytorch=2.0.1
    \item pytorch-cuda=11.8
    \item scikit-learn=1.2.2
    \item scipy=1.10.1
    \item seaborn=0.12.2
    \item shap-hypetune=0.2.6
    \item shap=0.41.0
    \item torchvision=0.15.2
    \item tqdm=4.65.0
    \item wandb=0.15.4
    \item xgboost=1.7.5
\end{itemize}

\section*{Supplementary Note 3 -- Additional Results}
In this section, we provide additional details of performance comparisons, distribution of severity scores across demographic groups, the impact of different feature selectors on the best model, hyper-parameter search space for the best model, and test statistics of how MediaPipe confidence varied after injecting different levels of external noise.

\begin{table*}[t]
\centering
\resizebox{1.95\columnwidth}{!}{%
\begin{tabular}{|l|l|l|l|l|l|l|l|l|l|}
\hline
\textbf{Comparison Groups} & \textbf{MAE} & \textbf{MSE} & \textbf{Accuracy} & \textbf{Kendall's Tau} & \textbf{MAPE} & \textbf{PCC} & \textbf{Spearman's $\rho$} \\ \hline
Expert vs Ground Truth     &     0.265         & 0.286             &     74.5\%          &    0.816                   &        12.79\%       &      0.863        &         0.862              \\ \hline
Expert vs Expert           &     \textbf{0.531}         &  \underline{0.623}            & \textbf{51.35\% }             &  \textbf{0.65}                     &  \textbf{25.78\% }            &  \textbf{0.722}            &   \textbf{0.717}                    \\ \hline
%Non-Expert 1 vs Expert       &   0.828           &      1.294        &       38.12\%        &     0.439                  &      63.15\%         &    0.509          &       0.495               \\ \hline
%Non-Expert 2 vs Expert       &    0.919          &     1.366         &     28.72\%          &   \underline{0.565}                    &       29.32\%        &      0.634        &          0.632             \\ \hline
Non-Expert vs Expert       &      0.873        &      1.33        &       33.42\%        &    0.502                   &     46.2\%          &     0.571         &      0.564                 \\ \hline
Non-Expert vs Ground Truth &     0.826         &     1.233         &      36.03\%         &      \underline{0.534}                 &        44.06\%       &       0.609       &           0.598            \\ \hline
Best Model vs Ground Truth &     \underline{0.580}         &        \textbf{0.536}      &      \underline{50.92\%}         &     0.515                  &        \underline{32.01\%}       &       \underline{0.656}       &          \underline{0.643}             \\ \hline
\end{tabular}
}
\caption{\textbf{Performance of experts, non-experts, and the model.} ``Expert vs Ground Truth'' assesses the average agreement between an expert's ratings and the ground-truth severity scores. ``Expert vs Expert'' measures the consistency of ratings among pairs of experts. ``Non-Expert vs Expert'' examines the association between ratings from non-experts and experts on average. ``Non-Expert vs Ground Truth'' evaluates the performance of non-experts by comparing their ratings to the ground-truth scores. Lastly, ``Best Model vs Ground Truth'' measures the performance of the best model against the ground-truth scores. Considering that the ground-truth scores are derived from the experts' ratings, it is expected that ``Expert vs Ground Truth'' would exhibit the strongest association. Among the other comparison groups, the best metrics are highlighted as \textbf{bold}, while the second best metrics are \underline{underlined}.}
\label{tab:performance_comparisons}
\end{table*}

% \usepackage{multirow}
\begin{table*}[]
\centering
\resizebox{1.95\columnwidth}{!}{%
\begin{tabular}{|llllllll|}
\hline
\multicolumn{2}{|c|}{\multirow{2}{*}{\textbf{Characteristics}}} & \multicolumn{5}{c|}{\textbf{Severity Score}}                                                                                                                                  & \multicolumn{1}{c|}{\multirow{2}{*}{\textbf{Total}}} \\ \cline{3-7}
\multicolumn{2}{|c|}{}                                          & \multicolumn{1}{l|}{\textbf{0}}  & \multicolumn{1}{l|}{\textbf{1}}   & \multicolumn{1}{l|}{\textbf{2}}   & \multicolumn{1}{l|}{\textbf{3}}  & \multicolumn{1}{l|}{\textbf{4}} & \multicolumn{1}{c|}{}                                \\ \hline
\multicolumn{2}{|l|}{\textit{Number of videos, n (\%)}}                       & \multicolumn{1}{l|}{108 (22.1\%)}         & \multicolumn{1}{l|}{181 (37.0\%)}          & \multicolumn{1}{l|}{141 (28.8\%)}          & \multicolumn{1}{l|}{54 (11.0\%)}          & \multicolumn{1}{l|}{5 (1.0\%)}          & 489 (100\%)                                                  \\ \hline
\multicolumn{8}{|l|}{\textit{Sex, n (\%)}}                                                                                                                                                                                                                                                                      \\ \hline
                & \multicolumn{1}{l|}{Male}                     & \multicolumn{1}{l|}{48 (44.4\%)} & \multicolumn{1}{l|}{100 (55.2\%)} & \multicolumn{1}{l|}{80 (56.7\%)}  & \multicolumn{1}{l|}{35 (64.8\%)} & \multicolumn{1}{l|}{4 (80\%)}   & 267 (54.6\%)                                         \\ \hline
                & \multicolumn{1}{l|}{Female}                   & \multicolumn{1}{l|}{60 (55.6\%)} & \multicolumn{1}{l|}{81 (44.8\%)}  & \multicolumn{1}{l|}{61 (43.3\%)}  & \multicolumn{1}{l|}{19 (35.2\%)} & \multicolumn{1}{l|}{1 (20\%)}   & 222 (45.4\%)                                         \\ \hline
\multicolumn{8}{|l|}{\textit{Race, n (\%)}}                                                                                                                                                                                                                                                                     \\ \hline
                & \multicolumn{1}{l|}{White}                    & \multicolumn{1}{l|}{91 (84.3\%)} & \multicolumn{1}{l|}{169 (93.4\%)} & \multicolumn{1}{l|}{136 (96.5\%)} & \multicolumn{1}{l|}{51 (94.4\%)} & \multicolumn{1}{l|}{5 (100\%)}  & 452 (92.4\%)                                         \\ \hline
                & \multicolumn{1}{l|}{Non-white}                & \multicolumn{1}{l|}{17 (15.7\%)} & \multicolumn{1}{l|}{12 (6.6\%)}   & \multicolumn{1}{l|}{5 (3.5\%)}    & \multicolumn{1}{l|}{3 (5.6\%)}   & \multicolumn{1}{l|}{0 (0\%)}    & 37 (7.6\%)                                           \\ \hline
\multicolumn{8}{|l|}{\textit{Recording environment, n (\%)}}                                                                                                                                                                                                                                                    \\ \hline
                & \multicolumn{1}{l|}{Home}                     & \multicolumn{1}{l|}{87 (80.6\%)} & \multicolumn{1}{l|}{136 (75.1\%)} & \multicolumn{1}{l|}{119 (84.4\%)} & \multicolumn{1}{l|}{46 (85.2\%)} & \multicolumn{1}{l|}{5 (100\%)}  & 393 (80.4\%)                                         \\ \hline
                & \multicolumn{1}{l|}{Clinic}                   & \multicolumn{1}{l|}{21 (19.4\%)} & \multicolumn{1}{l|}{45 (24.9\%)}  & \multicolumn{1}{l|}{22 (15.6\%)}  & \multicolumn{1}{l|}{8 (14.8\%)}  & \multicolumn{1}{l|}{0 (0\%)}    & 96 (19.6\%)                                          \\ \hline
\end{tabular}%
}
\caption{\textbf{Distribution of severity scores across demographic groups.}} 
\label{tab:severity_demography}
\end{table*}

% \usepackage{multirow}
\begin{table*}[]
\centering
\resizebox{1.95\columnwidth}{!}{%
\begin{tabular}{|c|l|l|l|l|l|l|l|l|}
\hline
\multicolumn{1}{|l|}{\textbf{\begin{tabular}[c]{@{}l@{}}Feature selection\\ method\end{tabular}}} & \textbf{Base model} & \textbf{MAE}    & \textbf{MSE}    & \textbf{Accuracy} & \textbf{Kendal's Tau} & \textbf{MAPE}    & \textbf{PCC}    & \textbf{Spearman's r} \\ \hline
\multirow{2}{*}{BoostRFE}                                                                         & LightGBM            & \textbf{0.5802} & 0.5364          & 50.92\%           & \textbf{0.5147}       & \textbf{32.01\%} & \textbf{0.6563} & \textbf{0.6429}       \\ \cline{2-9} 
                                                                                                  & XGBoost             & 0.5855          & \textbf{0.5251} & \textbf{51.94\%}  & 0.5032                & 33.41\%          & 0.6446          & 0.6309                \\ \hline
\multirow{2}{*}{BoostRFA}                                                                         & LightGBM            & 0.5993          & 0.5679          & 50.92\%           & 0.4911                & 33.56\%          & 0.63            & 0.6149                \\ \cline{2-9} 
                                                                                                  & XGBoost             & 0.5861          & 0.5586          & \textbf{51.94\%}  & 0.5044                & 32.14\%          & 0.6388          & 0.6329                \\ \hline
\end{tabular}%
}
\caption{\textbf{Effect of different feature selectors.} The performance of the best model (LightGBM regressor) is reported with different feature selectors we experimented with. The best metrics are highlighted as \textbf{bold}.} 
\label{tab:feature_selectors}
\end{table*}

\begin{table*}[]
\centering
\resizebox{1.95\columnwidth}{!}{%
\begin{tabular}{|l|l|l|l|}
\hline
\textbf{Name of the hyper-parameter}                                       & \textbf{Distribution} & \textbf{Values/Range}                                                          & \textbf{Best Value} \\ \hline
learning rate                                                              & Uniform               & {[}0.01, 0.3{]}                                                                & 0.01313             \\ \hline
\begin{tabular}[c]{@{}l@{}}maximum depth \\ of decision tree\end{tabular}  & Uniform               & {[}3, 18{]}                                                                    & 3                   \\ \hline
number of estimators                                                       & Uniform               & {[}25, 1000{]}                                                                 & 611                 \\ \hline
random state                                                               & Uniform               & {[}0, 8192{]}                                                                  & 42                  \\ \hline
seed                                                                       & Uniform               & {[}0, 8192{]}                                                                  & 42                  \\ \hline
use feature selection?                                                     & N/A                   & \{``yes'', ``no''\}                                                                & ``yes''               \\ \hline
method of feature selection                                                & N/A                   & \begin{tabular}[c]{@{}l@{}}\{``BoostRFE'', \\ ``BoostRFA''\}\end{tabular}          & ``BoostRFE''          \\ \hline
\begin{tabular}[c]{@{}l@{}}base model for \\ feature selector\end{tabular} & N/A                   & \begin{tabular}[c]{@{}l@{}}\{``XGBoost'', \\ ``LightGBM''\}\end{tabular}           & ``LightGBM''          \\ \hline
number of top features                                                     & Uniform               & {[}2, 53{]}                                                                    & 22                  \\ \hline
subsample                                                                  & Uniform               & {[}0.1, 1{]}                                                                   & 0.8                 \\ \hline
use feature scaling?                                                       & N/A                   & \{``yes'', ``no''\}                                                                & ``yes''               \\ \hline
method of feature scaling                                                  & N/A                   & \begin{tabular}[c]{@{}l@{}}\{``StandardScaler'',\\ ``MinMaxScaler''\}\end{tabular} & ``StandardScaler''    \\ \hline
\begin{tabular}[c]{@{}l@{}}use minority oversample\\ (SMOTE)\end{tabular}  & N/A                   & \{``yes'', ``no''\}                                                                & ``no''                \\ \hline
\end{tabular}%
}
\caption{\textbf{Hyper-parameter search space for the best model (LightGBM).} For hyper-parameters with a specific set of pre-defined values, the distribution is not applicable (N/A).} 
\label{tab:hyperparameters}
\end{table*}

\begin{table*}[t]
\centering
\resizebox{1.95\columnwidth}{!}{%
    \begin{tabular}{|l|l|l|l|l|}
\hline
\textbf{Comparison groups}                                                                   & \textbf{\begin{tabular}[c]{@{}l@{}}mean (std) for\\ the first group\end{tabular}} & \textbf{\begin{tabular}[c]{@{}l@{}}mean (std) for\\ the second group\end{tabular}} & \textbf{test-statistic} & \textbf{p-value}        \\ \hline
\begin{tabular}[c]{@{}l@{}}Original vs \\ slightly blurred videos\end{tabular}               & 0.96 (0.031)                                                                      & 0.958 (0.033)                                                                      & 0.289                   & 0.773                   \\ \hline
\begin{tabular}[c]{@{}l@{}}Original vs \\ substantially blurred videos$^*$\end{tabular}          & 0.96 (0.031)                                                                      & 0.935 (0.08)                                                                       & 2.624                   & 0.009                   \\ \hline
\begin{tabular}[c]{@{}l@{}}Slightly blurred vs \\ substantially blurred videos$^*$\end{tabular}  & 0.958 (0.033)                                                                     & 0.935 (0.08)                                                                       & 2.445                   & 0.016                   \\ \hline
\begin{tabular}[c]{@{}l@{}}Original vs \\ videos with small added noise$^*$\end{tabular}         & 0.96 (0.031)                                                                      & 0.897 (0.124)                                                                      & 4.530                   & $1 \times 10^{-5}$  \\ \hline
\begin{tabular}[c]{@{}l@{}}Original \\ vs videos with high added noise$^*$\end{tabular}          & 0.96 (0.031)                                                                      & 0.784 (0.170)                                                                      & 9.364                   & $5 \times 10^{-17}$ \\ \hline
\begin{tabular}[c]{@{}l@{}}Videos with small added noise vs \\ high added noise$^*$\end{tabular} & 0.897 (0.124)                                                                     & 0.784 (0.170)                                                                      & 4.944                   & $2 \times 10^{-6}$  \\ \hline
\end{tabular}%
}
\caption{\textbf{Test statistics.} Details of the statistical tests for assessing the difference in MediaPipe confidence scores across video groups of different quality. We used a two-tailed paired sample t-test to report the statistical significance. The comparison groups with a star$^*$ demonstrate statistically significant differences (at level of significance $\alpha = 0.05$).}
\label{tab:test_statistics}

\end{table*}